\documentclass[conference]{IEEEtran}
\IEEEoverridecommandlockouts
\usepackage{cite}
\usepackage{amsmath,amssymb,amsfonts}
\usepackage{graphicx}
\usepackage{textcomp}
\usepackage{xcolor}
\def\BibTeX{{\rm B\kern-.05em{\sc i\kern-.025em b}\kern-.08em
    T\kern-.1667em\lower.7ex\hbox{E}\kern-.125emX}}
\usepackage{comment}
\usepackage{tikz} 
\usetikzlibrary{calc,arrows.meta,positioning,shapes.geometric, arrows}
\tikzstyle{process} = [rectangle, minimum width=3cm, minimum height=1cm, text centered, text width=4cm, draw=black, fill=orange!30]
\usepackage{dsbda}
\usepackage{datetime}
\usepackage{caption}
\usepackage{subcaption}
\usepackage{tabularx}
\usepackage{listings}
\usepackage{svg}
\usepackage{algorithm}
\usepackage[]{algpseudocode} 
\algtext*{EndProcedure}
\usepackage{pgf}
\usepackage[utf8]{inputenc}\DeclareUnicodeCharacter{2212}{-}
\usepackage{import}

\definecolor{gold}{rgb}{0.85, 0.65, 0.13}

\newcommand{\todoorange}[1]{\todo[color=orange,inline]{#1}}

\newcommand{\AttributeCollection}{Attribute Collection}
\newcommand{\ClassCollection}{Class Collection}
\newcommand{\PropertyTypeCollection}{Property Type Collection}
\newcommand{\Schemex}{SchemEX}

\newcommand{\Mi}[1]{$M_\text{#1}$}
\newcommand{\mattrcol}{\Mi{AC}}
\newcommand{\mclasscol}{\Mi{CC}}
\newcommand{\mpropertytypecol}{\Mi{PTC}}
\newcommand{\mschemex}{\Mi{SX}}

\newcommand{\mattrcolns}{\Mi{AC-NS}}
\newcommand{\mclasscolns}{\Mi{CC-NS}}
\newcommand{\mpropertytypecolns}{\Mi{PTC-NS}}

\newcommand{\dyldosmall}{DyLDO-25\%}
\newcommand{\mattrcolsmall}{\Mi{AC-25}}
\newcommand{\mclasscolsmall}{\Mi{CC-25}}
\newcommand{\mpropertytypecolsmall}{\Mi{PTC-25}}

\newcommand{\dyldoschemex}{DyLDO-\textit{6\_499}}
\newcommand{\mschemexused}{\Mi{SX-\textit{6\_499}}}

\newcommand{\dyldothirty}{DyLDO-30M}
\newcommand{\mattrcolthirty}{\Mi{AC-30M}}
\newcommand{\mclasscolthirty}{\Mi{CC-30M}}
\newcommand{\mpropertytypecolthirty}{\Mi{PTC-30M}}

\definecolor{darkgreen}{rgb}{0,0.75,0}
\newcommand{\topOne}[1]{\textbf{\color{darkgreen}#1}}
\newcommand{\topTwo}[1]{\textbf{\color{blue}#1}}
\newcommand{\topThree}[1]{\textbf{\color{red}#1}}
\newcommand{\topFour}[1]{\textbf{\color{orange}#1}}

\newcommand{\GraphMLP}{Graph-MLP\xspace}

\def\mTheta{{\bm{\Theta}}}
\newcommand{\relu}{\mathrm{ReLU}}
\newcommand{\mean}{\mathrm{MEAN}}

\AtBeginDocument{
  \providecommand\BibTeX{{
    \normalfont B\kern-0.5em{\scshape i\kern-0.25em b}\kern-0.8em\TeX}}}

        \hypersetup{
           breaklinks=true,   
           colorlinks=true,   
           pdfusetitle=true,  
        }
        
\begin{document}

\title{Graph Summarization as Vertex Classification Task using Graph Neural Networks vs. Bloom Filter}

\author{\IEEEauthorblockN{M. Blasi, M. Freudenreich, J. Horvath}
\IEEEauthorblockA{
\textit{Ulm University}, Germany \\
\{maximilian.blasi,manuel.freudenreich,\\
johannes.horvath\}@uni-ulm.de
}
\and
\IEEEauthorblockN{D. Richerby}
\IEEEauthorblockA{\textit{University of Essex, UK} \\
david.richerby@essex.ac.uk}
\and
\IEEEauthorblockN{A. Scherp}
\IEEEauthorblockA{
\textit{Ulm University}, Germany \\
ansgar.scherp@uni-ulm.de}
}

\maketitle

\begin{abstract}
\shortorextended{The goal of graph summarization is to represent large graphs in a structured and compact way. A graph summary based on equivalence classes preserves predefined features of each vertex within a $k$-hop neighborhood, such as the vertex and edge labels. Based on these neighborhood characteristics, the vertex is assigned to an equivalence class. The calculation of the assigned equivalence class must be a permutation invariant operation on the predefined features. This is typically achieved by sorting on the feature values, which is computationally expensive, and subsequently hashing the result. Graph Neural Networks (GNNs) fulfill the permutation invariance requirement. We formulate the problem of graph summarization as a subgraph classification task on the root vertex of the $k$-hop neighborhood. We adapt different GNN architectures, both based on the popular message-passing protocol and alternative approaches, to perform the structural graph summarization task. We compare different GNNs with a standard multi-layer perceptron (MLP) and Bloom filter as a non-neural method. We consider four popular graph summary models on a large web graph. This resembles challenging multi-class vertex classification tasks with the numbers of classes ranging from 576 to hundreds of thousands. Our results show that the performance of GNNs are close to each other. In three out of four experiments, the non-message-passing \GraphMLP model outperforms the other GNNs. The performance of the standard MLP is extraordinarily good, especially in the presence of many classes. Finally, the Bloom filter outperforms all neural architectures by a large margin, except for the dataset with the fewest number (576) of classes.}
{The goal of graph summarization is to represent large graphs in a structured and compact way. 
A graph summary based on equivalence classes preserves predefined features of a graph's vertex within a $k$-hop neighborhood such as the vertex labels and edge labels.
Based on these neighborhood characteristics, the vertex is assigned to an equivalence class. 
The calculation of the assigned equivalence class must be a permutation invariant operation on the predefined features. 
This is achieved by sorting on the feature values, e.\,g., the edge labels, which is computationally expensive, and subsequently hashing the result. 
Graph Neural Networks (GNNs) fulfill the permutation invariance requirement.
We formulate the problem of graph summarization as a subgraph classification task on the root vertex of the $k$-hop neighborhood.
We adapt different GNN architectures, both based on the popular message-passing protocol and alternative approaches, to perform the structural graph summarization task.
We compare different GNNs with a standard multi-layer perceptron (MLP) and Bloom filter as non-neural method. 
For our experiments, we consider four popular graph summary models on a large web graph.
This resembles challenging multi-class vertex classification tasks with the numbers of classes ranging from $576$ to multiple hundreds of thousands. 
Our results show that the performance of GNNs are close to each other.
In three out of four experiments, the non-message-passing \GraphMLP model outperforms the other GNNs. 
The performance of a standard-MLP baseline is on par with the GNNs. 
However, the Bloom filter outperforms all neural architectures by a large margin, except for the dataset with the fewest number of $576$ classes.}
This is an interesting result, since it sheds light on how well and in which contexts GNNs are suited for graph summarization.
Furthermore, it demonstrates the need for considering strong non-neural baselines for standard GNN tasks such as vertex classification.

Our source code is available at
\url{https://github.com/johorvath/Graph_Summarization_with_Graph_Neural_Networks}

\end{abstract}


\section{Introduction}
\label{sec:introduction}

Graph summaries provide a condensed representation of an input graph.
The goal is to preserve predefined features that are relevant for specific tasks such as queries on the summary~\cite{BLUME2021136}. 
Graph summaries are used for tasks such as estimating cardinalities~\cite{DBLP:conf/icde/NeumannM11}; data search~\cite{DBLP:journals/ws/KonrathGSS12}, exploration~\cite{7396819}, and visualization~\cite{DBLP:journals/vldb/GoasdoueGM20}; and vocabulary term recommendations~\cite{10.1007/978-3-319-34129-3_7}.
Lossless summaries preserve all information needed for the desired task, allowing it to be computed exactly from the summary.
In this paper, we consider graph summaries based on equivalence classes~\cite{BLUME2021136}.
These classify vertices based on features such as vertices labels, edges, and neighbors to produce a lossless summary.
With the recent rise of graph neural networks (GNNs), our contribution is to understand how well and in which context GNNs are suited to computing
approximations of lossless
graph summaries, and how they compare to Bloom filters, as a non-neural baseline.

\begin{figure}[htb]
    \centering
    \begin{subfigure}{.4\textwidth}
        \centering
        \begin{minipage}{0.4\textwidth}
            ($x_1$,$p_1$,$y_1$)
            
            ($x_1$,$p_2$,$y_2$)
            \newline
            
            $List_{Hash}$: [$p_1$,$p_2$]
        \end{minipage}
        \begin{minipage}{0.4\textwidth}
            \begin{tikzpicture}[node distance={15mm}, thick, main/.style = {draw, circle}]
            \node[main] (1) {$x_1$};  
            \node[main] [above right of=1] (2) {$y_1$};
            \node[main] [below right of=1] (3) {$y_2$}; 
            \draw[->] (1) -- node[midway, above, sloped, pos=0.5] {$p_1$} (2);
            \draw[->] (1) -- node[midway, below, sloped, pos=0.5] {$p_2$} (3);
            \end{tikzpicture} 
    \vspace{3mm}
        \end{minipage}
    \end{subfigure}
    \begin{subfigure}{.4\textwidth}
        \centering
        \begin{minipage}{0.4\textwidth}
            ($x_2$,$p_2$,$y_2$)
            
            ($x_2$,$p_1$,$y_1$)
            \newline
            
            $List_{Hash}$: [$p_2$,$p_1$]
        \end{minipage}
        \begin{minipage}{0.4\textwidth}
            \begin{tikzpicture}[node distance={15mm}, thick, main/.style = {draw, circle}]
            \node[main] (1) {$x_2$};  
            \node[main] [above right of=1] (2) {$y_2$};
            \node[main] [below right of=1] (3) {$y_1$}; 
            \draw[->] (1) -- node[midway, above, sloped, pos=0.5] {$p_2$} (2);
            \draw[->] (1) -- node[midway, below, sloped, pos=0.5] {$p_1$} (3);
            \end{tikzpicture} 
        \end{minipage}
    \end{subfigure}
    \caption{The left-hand side shows two list-based representations of the two graphs on the right-hand side. 
    While the two graphs are actually identical, the order of triples in the lists differ.
    In a naive approach of hashing the corresponding lists of triples, different hash values and thus different equivalence classes of the vertex $x_1$ and $x_2$ are computed.
    However, the two subgraphs should yield the same equivalence class for $x_1$ and $x_2$ based on the graph summary model \AttributeCollection\ (see \autoref{subsec:graphSummaryM}) used in this example. }
    \label{fig:example-hash}
\end{figure}
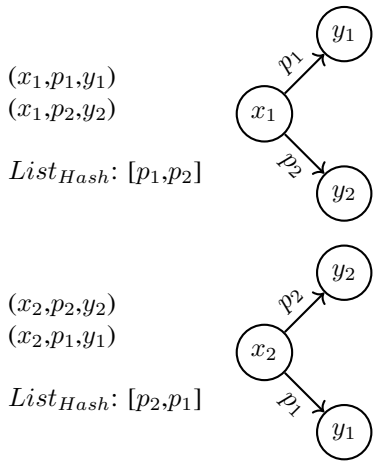

In general, graph summarization can be understood as a function $f_{M,G}$ that assigns each vertex in the graph $G$ to an equivalence class:
\begin{equation}
f_{M,G}\colon V(G)\mapsto V(SG)\,,
\end{equation}
where $V(G)$ is the vertex set of~$G$, and $SG$ is the summary graph, whose vertices correspond to equivalence classes.
The content of these equivalence classes is determined by the graph summary model \Mi{}. 
Each model \Mi{} considers different properties of neighboring vertices for the calculation. 
So the summary function maps a set of information gathered in the subgraph around a vertex $v$ to a concise representation corresponding to the equivalence class.
The aggregation of the information from a vertex's subgraph is required to be isomorphism-invariant, \ie it must depend only on the structure of the graph itself, and not on, \eg the order in which the edges are listed in the database.
\autoref{fig:example-hash} illustrates this requirement for isomorphism-invariance.

The calculation of the equivalence class can be done using deterministic computations like a hash function~\cite{DBLP:journals/ws/KonrathGSS12}. 
Because the hash value resulting from $f_{M,G}$ corresponds to an equivalence class, we can interpret it as a vertex label and apply a neural network such as a GNN for the classification. 
In other words, we cast the problem of graph summarization as a vertex classification task in GNNs.
This is possible because a key design feature for GNNs is that they should be permutation invariant or at least equivariant~\cite{hamilton2020graph}, as required for this application.
The use of neural networks as hash functions has so far mainly been researched in the context of security~\cite{turvcanik2016hash, lian2007one}. 
There, the goal is to provide a one-way function that ensures big changes in the resulting hash, even when there are only small differences in the input. We discuss this in more detail in \autoref{subsec:NNAsHashRW}.

We consider different GNN models, both based on the popular message-passing architecture, namely the 
graph convolutional network (GCN)~\cite{kipf2016semi}, GraphSAGE~\cite{hamilton2018inductive}, and GraphSAINT~\cite{DBLP:journals/corr/abs-1907-04931}, and an approach that does not use message-passing, namely \GraphMLP~\cite{graphMLP}.
GCN applies a message-passing technique for vertex classification, based on the vertex embedding and edges. 
GraphSAGE applies functions with trainable weights for vertex classification instead of trainable weights directly on the vertex embeddings. 
GraphSAINT uses a sampler for batching and then a GCN defined on the dimension of the smaller batch. 
\GraphMLP removes the message passing technique and replaces it with a specific loss function and multi-layer perceptron (MLP) network.
To apply the GNNs on large-scale datasets and avoid the general memory limitations of GNNs, we implemented a vertex sampler to generate mini-batches. 
This is similar to the NodeSampler used in GraphSAINT, but our sampler takes the class distribution into account by using a distribution inverse to the number of occurrences of classes. 
We use a vanilla MLP as baseline model on a binary encoded feature matrix (one-hot encoded labels). 
We compare a Bloom filter approach and measure its performance when computing graph summaries. 
The Bloom filter is used as a non-neural baseline. 
A specific characteristic of the filter is that it does not require the relevant feature values to be sorted for summarizing vertices. 
Overall, we evaluate six different methods: four GNNs, one MLP, and the Bloom filter.

We apply four different popular graph summary models~\cite{BLUME2021136} based on features that consider the edge labels (see \autoref{fig:example-hash}), vertex labels, their combination (edge labels plus vertex labels), as well as a model based on a the labels of a vertex's neighbors.
We compute four graph summaries using these models for the DyLDO dataset~\cite{kafer2012towards}, which is a subset of linked data  crawled from a set of seed-URIs. 
We randomly split the DyLDO graph into training and test sets to evaluate different classifier models on the four summaries. 
The DyLDO dataset was chosen because the data is stored in a standardized way and because it is representative of real-world use of graphs on the web, but its size is still manageable.

In summary, our research questions are:
\begin{itemize}
    \item Can we produce a graph summary via a vertex classification based on GNNs? 
    How accurate is a graph summary calculated by a GNN?
    \item How do different classifiers perform against each other at calculating a graph summary? How do more sophisticated models perform against a baseline?
    \item Do the classifiers perform differently depending on the summary model \Mi{}?
\end{itemize}

The GNNs are evaluated by a $10$-fold cross-validation, following the recommendation of~\cite{pitfalls}. 
In our experiments, the Bloom filter outperforms all GNNs and the MLP baseline. 
The results for the GNNs are close to each other, and in three out of four graph summary models the non-message-passing \GraphMLP outperforms the message-passing GNNs. 
The performance of the standard MLP is close to GNNs and ranked third or fourth in three of the experiments.
Particularly in the presence of many classes, the standard MLP is on par with the GNNs.
We run an ablation study on the DyLDO dataset, where singleton class occurrences are removed, reducing the complexity of the classification task.
The experiments show that the results improved but interestingly not for all models. 
Further ablation studies are reported in our extended report~\cite{DBLP:journals/corr/abs-2203-05919}.
First, we applied $1$-hop GNN models on $1$-hop graph summaries, which reduced the standard error but overall did not improve the results.
Finally, we investigated if widening the MLP's hidden layer, \ie increasing its capacity, improves the result.
Doubling the number of hidden nodes to $2{\small,}048$ slightly improves the results but cannot reach the performance of the other models.

Below, we discuss the related works.
We describe models we use to answer the research questions in \autoref{sec:models}. 
The experimental apparatus, \autoref{sec:expApparatus}, contains an introduction into the dataset, experimental procedure, implementation, and evaluation metrics. 
The results of our experiments are reported in \autoref{sec:results}. 
We discuss the results in \autoref{sec:discussion}, before we conclude. Note that we use the term \textit{node} for the use in the neural network context and \textit{vertex} in the graph context.

\section{Related Work}
\label{sec:relWork}
First, we summarize the state of the art for graph summarization.
Thereafter, we review different neural network models suited for the vertex classification task.
We continue discussing hash functions in the context of neural networks.
Finally, an introduction to Bloom filters is given. 

\subsection{Graph Summarization}
\label{subsec:graphSummarizationRW}
Graph summaries are a good approach to provide a structural representation of large graph datasets~\cite{BLUME2021136}.
Below, we define our notion of graphs and the task of graph summarization.

\paragraph{Graph}
A graph $G=(V,E,R)$ consists of a set~$V$ of vertices, a set~$R$ of relation types and a set $E\subseteq V\times R\times V$ of labeled edges. Vertices typically represent entities, and edges describe the relationships between them.

\paragraph{RDF Graphs and \texttt{rdf:type}}
The Resource Description Framework (RDF) is a W3C standard that models graphs as \textit{subject--predicate--object} triples $(s, p, o)$~\cite{N-Triple}. For our purposes, we assume that $s,o\in V$, $p\in R$, and $(s,p,o)\in E$.
\extended{It can also be expanded from triples to quads, where an optional fourth value describing the \textit{graph} in the dataset the triple belongs to, is added to the tuple~\cite{N-Quads}.}  
A special predicate in the RDF standard is \texttt{rdf:type}. The triple ($s_i$, \texttt{rdf:type}, $o_j$) denotes that $s_i$~has the vertex label $o_j$. Predicates $p \neq \texttt{rdf:type}$ are called RDF properties.

\paragraph{Graph Summary}
As discussed above, the idea of graph summarization is to generate a condensed representation $SG$ of an input graph $G$~\cite{BLUME2021136}. 
The result is a summarized graph which is typically much smaller than~$G$ but which preserves structural information necessary for a given task.
This compression allows tasks to be computed on the graph summary much faster than on the original graph~\cite{BLUME2021136}, \eg  counting vertices that have the same information in the neighborhood.
To this end, information about each vertex's neighborhood in the graph~$G$ is gathered and combined in a specific way defined by the summary model~\Mi{}.
Vertices are classed as equivalent based on the properties defined by~\Mi{}.
For example, the Attribute Collection summary model determines vertices to be equivalent if they have the same set of RDF properties~\cite{BLUME2021136}.
The summary graph~$SG$ has one vertex per equivalence class and edges between classes corresponding to the neighborhood structure identified by the summary model.

The neighborhood features extracted by the summary model can be ``compressed'' by a hash function to give a class label for the equivalence class.
This labeling can be seen as a vertex classification task, and we investigate the extent to which this task can be performed by neural networks.

\extended{
\paragraph{FLUID}
A good way to calculate graph summaries is to use FLUID~\cite{BLUME2021136}. It is a language and generic algorithm for flexibly defining and adapting graph summaries and is able to define all existing lossless structural graph summaries. However the computation time increases for larger graphs when using the FLUID framework because of the usage of sorting and hashing operations. The runtime of those operations depends on the graph size, resulting in long computation times for large graphs.
}

\subsection{Graph Neural Networks for Vertex Classification}
\label{subsec:gnnRW}

For the following section, we denote by $\mA$ an adjacency matrix and by $\mD$ the corresponding degree matrix. The adjacency matrix with self loops is calculated by $\tilde\mA = \mA + \mI_N$ with the identity matrix $\mI_N$. $\tilde\mD$ corresponds to the degree matrix of $\tilde\mA$.

\paragraph{GCN and Variants}
To run a classification task on graphs, Kipf et al.~\cite{kipf2016semi} suggest a graph convolutional network (GCN) model
\begin{equation}
\label{eqn:gcn}
    f_\text{GCN}(\mX,\mA) = \softmax \big( \hat{\mA} \cdot \relu \big(\hat{\mA} \cdot ( \mX \mW^{(0)} ) \big) \cdot \mW^{(1)} \big)
\end{equation}
applied directly on the graph~$G$ using message-passing. 
Here, $\mX$ denotes a matrix containing vectors of node features, $\hat{\mA}= \tilde\mD ^{-1/2}\tilde\mA\tilde\mD ^{-1/2}$ is the symmetric Laplacian of $\tilde\mA$ (symmetrically normalized by vertex degree), and $\mW^{(i)}$ denotes the weight matrix of the hidden layer $i$.
For message-passing, each vertex generates a message based on its embedding and sends this message to all its neighbors. Each vertex then aggregates the received messages in a permutation invariant manner and updates its own embedding with the aggregation result.
To handle semi-supervised training data, the function $f_\text{GCN}(\mX,\mA)$ is conditioned through the loss function on the vertices that have labels. But during training and inference there are also vertices present that do not have any label.
Kipf et al.~\cite{kipf2016semi} show that a two-layer GCN results in a $2$-hop aggregation. 

Schlichtkrull et al.~\cite{schlichtkrull2017modeling} propose relational GCN (R-GCN) that extends the GCN model by defining a directed graph $G=(V,E,R)$. The model adds a trainable weight matrix for every relation type $r \in R$. Now a GCN can also be applied for entity classification and link prediction on relational data. 
Another approach is Simple Graph Convolution (SGC) proposed by Wu et al.~\cite{DBLP:journals/corr/abs-1902-07153}. Their goal is to reduce the complexity of GCN by removing non-linearities and by condensing the weight matrices of single layers into a combined one. 

\paragraph{Sampling-based GNNs}
GCNs are limited by their memory requirements.
The weight matrices $\mW$ and node feature matrices $\mX$ are dimensioned on the full adjacency matrix $\mA$. 
This means that the whole graph has to be loaded into memory. 
Thus, batching is not possible.
Furthermore, only a transductive training is possible. 
These shortcomings are mitigated by sampling-based models like GraphSAINT and GraphSAGE.

GraphSAINT~\cite{DBLP:journals/corr/abs-1907-04931} uses a sampler (vertex-, edge-, or random walk-based) to generate smaller batches that are applied to a classic GCN model. 
The GCN model is dimensioned to fit the sampled graph $G_s=(V_s,E_s)$ and so it is greatly reduced in size. The weights of the smaller model have to be transferred from the complete model and then later updated. The activation $h_{v}^{k}$ of each vertex $v$ in layer $k$ is calculated via \autoref{eqn:graphsaint}, using an activation function $\sigma$, the normalized and sampled adjacency matrix $\tilde{\mA_s}$, and the weight matrix $\mW_s$, containing all necessary weights for the sampled model.
\begin{equation}
\label{eqn:graphsaint}
    h_{v}^{k} = \sigma\bigg(\sum_{u\in \gV_s} \tilde{A_s}[v,u] (\mW_s^{k-1})^{T} h_{u}^{(k-1)}\bigg)\,.
\end{equation} 

GraphSAGE~\cite{hamilton2018inductive} aggregates the information in a $k$-neighborhood into a vertex embedding of the current root vertex of the neighborhood.
Its authors explored mean, pre-trained LSTM, and pooling aggregators, which are permutation invariant. Training those aggregators instead of the weights of the vertices' embeddings, like in GCN, allows an inductive training process.
The node activation $h_{v}^{k}$ for $v \in \gV$ and the neighborhood $\gN$ of $v$ in layer $k$ is calculated by
\begin{equation}
\label{eqn:graphsage}
    h_{v}^{k} = \sigma\Big(\mW \cdot \mean\big(\{h_{\rv}^{k-1}\} \cup \{h_{u}^{k-1}\mid u \in \gN(\rv)\}\big)\Big)\,.
\end{equation} 

There are further sampling-based models~\cite{chen2018stochastic,DBLP:journals/corr/abs-1809-05343}.
\extended{Chen et al.~\cite{chen2018stochastic} propose a StochasticGCN model. 
It uses a control variate-based sampler to reduce the receptive field of a node. An activation history is kept per node and then used in the stochastic training process.
An adaptive sampling method is proposed by Huan et al.~\cite{DBLP:journals/corr/abs-1809-05343} to increase the convergence speed of a model. They apply a layerwise sampling approach which results in a linear growth of nodes.}
We leave the consideration of these models for future work.

\paragraph{\GraphMLP}
Hu et al.~\cite{graphMLP} introduce an MLP-based GNN without the message-passing mechanism used in conventional GNNs. The model can be split into a two-layer MLP followed by a classifier layer. In the MLP, a GELU-activation function $\sigma$, layer-normalization, and dropout is applied (see Equations \ref{eqn:XgraphMLP} and \ref{eqn:ZgraphMLP}, where $\mX^{(i)}$ and $\mW^{(i)}$ denote the node feature matrix $\mX$ and the weight matrix $\mW$ of the $i$-th layer). 
Finally, a linear classifier layer is applied as shown in \autoref{eqn:YgraphMLP}.
\begin{subequations}
     \begin{equation}
        \label{eqn:XgraphMLP}
        \mX^{(1)}=\mathrm{Dropout}\big(LN\big(\sigma(\mX \mW^{(0)})\big)\big)
    \end{equation}
    \begin{equation}
        \label{eqn:ZgraphMLP}
        \mZ=\mX^{(1)}\mW^{(1)}
    \end{equation}
    \begin{equation}
        \label{eqn:YgraphMLP} 
        \mY = \mZ \mW^{(2)}
    \end{equation}
\end{subequations}

For feature transformation, a neighboring contrastive (NContrastive) loss $\textit{loss}_{NC}$ is introduced by applying it to the output layer of the MLP $\mZ$ (see \autoref{eqn:ZgraphMLP}). 
In the NContrastive loss, all vertices inside the $r$-hop neighborhood for each vertex are counted as positive samples and vertices outside this neighborhood as negative ones. The loss is calculated on the base of the $r$th power $\hat{\mA}^r$ of the normalized adjacency matrix and a cosine similarity with a temperature parameter $\tau$. 
To the output layer $\mY$ (see \autoref{eqn:YgraphMLP}), a standard cross-entropy loss $\textit{loss}_{CE}$ is applied for the classification objective. 
The NContrastive loss is weighted by coefficient $\alpha$ resulting in the $\textit{loss}_{total}=\textit{loss}_{CE}+\alpha \cdot \textit{loss}_{NC}$.

\subsection{Neural Networks as Hash Functions}
\label{subsec:NNAsHashRW}
The goal of a graph summary is the partitioning of the vertices into equivalence classes. Each class can be represented using a class label, which can be generated using a hash function.
Until now, the application of neural networks as a hash function has been considered mainly in the context of security.
Tur{\v{c}}an{\'\i}k et al.~\cite{turvcanik2016hash} initialize a neural network with random weights and biases.
Lian et al.~\cite{lian2007one} apply a chaotic map function to the weights and biases. 
The goal of those neural networks is to provide a one-way function and they aim to ensure that small changes in the input sequence create big changes in the resulting hash value. 

Our goal is not a general purpose hash function but a classifier. 
Therefore, the security aspects discussed in Lian et al.~\cite{lian2007one} are not relevant in our application. 
Nevertheless, since the result of the hash function can be interpreted as a class label for the graph summary's equivalence class, we can apply a neural network for this classification task.

\subsection{Bloom Filters} 
\label{subsec:BloomFilterRW} 
A Bloom filter is a probabilistic data structure used to test whether an element is a member of a set. False positive matches can occur with a low probability, but false negatives results are impossible~\cite{rottenstreich2014bloom}. A Bloom filter uses a Boolean array with a predefined size, each entry is initiated with \textit{FALSE}. When adding an element to the Bloom filter, the element is hashed by multiple predefined hash functions, resulting in multiple indices of the Bloom filter array and the values at these array indexes are then set to \textit{TRUE}. To query a membership in a Bloom filter, the element just needs to be hashed by the predefined hash functions and then checked whether the resulting array indices return \textit{TRUE}. If all of them do so, the element is most likely in the set, with a small chance of false positives, and if any of the returned values is \textit{FALSE}, the element is definitely not in the set~\cite{mitzenmacher2017probability}.

\section{Models}
\label{sec:models}

First, we explain how to compute graph summaries and provide an overview of the summary models considered in our work. 
Thereafter, we describe the graph neural networks used in our experiments.
Finally, we explain the multi-layer perceptron used as a neural baseline and Bloom filters as a non-neural baseline.

\subsection{Computing Graph Summaries with Neural Networks}
\label{subsec:graphSummaryM}

The main goal of this work is the calculation of graph summaries using neural network techniques. 
For training and evaluation, it is essential to determine the true class of each vertex. The class depends on the specified summary model $M$. 
The class labels are calculated in a ``traditional'', lossless way to establish a ground truth and to use them later as target labels for training the neural networks.

The information of vertex $v$ is defined through all the $(s, p, o)$-triples with $s=v$. The summary model defines which triples are considered for each subject. 
The vertex information is gathered in a list (see \autoref{fig:example-hash}) and sorted.
Based on that list, the equivalence class for each vertex is determined. 
In more detail, we collect the information as strings in a list, sort the list, concatenate the strings into a single string, and then apply a hash function. 
The resulting hash can then be used as the equivalence class. 
This approach is based on \Schemex{}~\cite{DBLP:journals/ws/KonrathGSS12}.

The sorting algorithm timsort~\cite{Heumann2021} is used, which is a combination of insertion sort and merge sort. 
For small input sizes, insertion sort is used with a complexity of $O(n^2)$ and for big sizes, merge sort with a complexity of $O(n\log{}n)$. This leads to timsort's worst-case and average complexity of $O(n\log{}n)$, and best-case complexity $O(n)$. For strings, timsort has the property that it uses fewer comparisons than other $\Theta(n\log n)$ sorting algorithms. This is advantageous for us, since it is our use case. 
We use the default hashing function for strings in Python, which is an implementation of the modified Fowler--Noll--Vo algorithm~\cite{Python2021}.

\subsection{Considered Graph Summary Models}
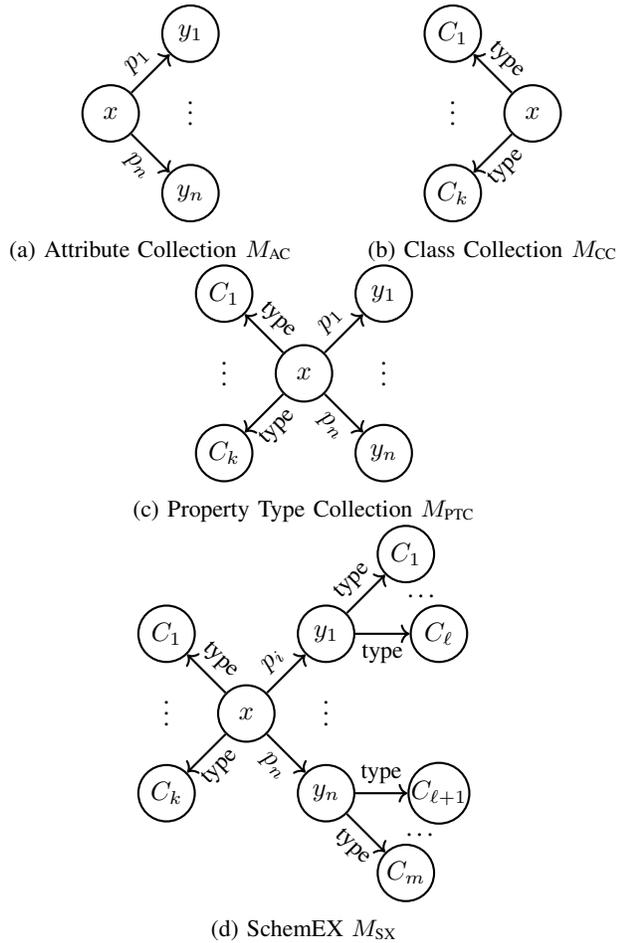
\begin{figure}[htb]
    \begin{subfigure}{.275\textwidth}
        \centering
        \begin{tikzpicture}[node distance={15mm}, thick, main/.style = {draw, circle,inner sep=0pt,minimum size=0.75cm}]
            \node[main] (1) {$x$};  
            \node[main] [above right of=1] (2) {$y_1$};
            \node[main] [below right of=1] (3) {$y_n$}; 
            \draw[->] (1) -- node[midway, above, sloped, pos=0.5] {$p_1$} (2);
            \draw[->] (1) -- node[midway, below, sloped, pos=0.5] {$p_n$} (3);
            
            \node at ($(2)!.45!(3)$) {\vdots};
        \end{tikzpicture} 
        \caption{Attribute Collection \mattrcol{}}
        \label{fig:collections_sub1}
    \end{subfigure}
    \begin{subfigure}{.2\textwidth}
        \centering
        \begin{tikzpicture}[node distance={15mm}, thick, main/.style = {draw, circle,inner sep=0pt,minimum size=0.75cm}]
            \node[main] (1) {$x$};  
            \node[main] [above left of=1] (2) {$C_1$};
            \node[main] [below left of=1] (3) {$C_k$}; 
            \draw[->] (1) -- node[midway, above, sloped, pos=0.5] {$\text{\small type}$} (2);
            \draw[->] (1) -- node[midway, below, sloped, pos=0.5] {$\text{\small type}$} (3);
            
            \node at ($(2)!.45!(3)$) {\vdots};
        \end{tikzpicture} 
        \caption{Class Collection \mclasscol{}}
        \label{fig:collections_sub2}
    \end{subfigure}
    \begin{subfigure}{.5\textwidth}
        \centering
        \begin{tikzpicture}[node distance={15mm}, thick, main/.style = {draw, circle,inner sep=0pt,minimum size=0.75cm}]
            \node[main] (1) {$x$};  
            \node[main] [above left of=1] (2) {$C_1$};
            \node[main] [below left of=1] (3) {$C_k$}; 
            \draw[->] (1) -- node[midway, above, sloped, pos=0.5] {$\text{\small type}$} (2);
            \draw[->] (1) -- node[midway, below, sloped, pos=0.5] {$\text{\small type}$} (3);
            \node[main] [above right of=1] (4) {$y_1$};
            \node[main] [below right of=1] (5) {$y_n$}; 
            \draw[->] (1) -- node[midway, above, sloped, pos=0.5] {$p_1$} (4);
            \draw[->] (1) -- node[midway, below, sloped, pos=0.5] {$p_n$} (5);
            
            \node at ($(2)!.45!(3)$) {\vdots};
            \node at ($(4)!.45!(5)$) {\vdots};
        \end{tikzpicture} 
        \caption{Property Type Collection \mpropertytypecol{}}
        \label{fig:collections_sub3}
    \end{subfigure}
    \begin{subfigure}{.5\textwidth}
        \centering
        \begin{tikzpicture}[node distance={15mm}, thick, main/.style = {draw, circle,inner sep=0pt,minimum size=0.75cm}]
            \node[main] (1) {$x$};  
            \node[main] [above left of=1] (2) {$C_1$};
            \node[main] [below left of=1] (3) {$C_k$}; 
            \draw[->] (1) -- node[midway, above, sloped, pos=0.5] {$\text{\small type}$} (2);
            \draw[->] (1) -- node[midway, below, sloped, pos=0.5] {$\text{\small type}$} (3);
            \node[main] [above right of=1] (4) {$y_1$};
            \node[main] [below right of=1] (5) {$y_n$}; 
            \draw[->] (1) -- node[midway, above, sloped, pos=0.5] {$p_i$} (4);
            \draw[->] (1) -- node[midway, below, sloped, pos=0.5] {$p_n$} (5);
            \node[main] [above right of=4] (6) {$C_1$};
            \node[main] [right of=4] (7) {$C_\ell$}; 
            \node[main] [right of=5] (8) {$C_{\ell+1}$};
            \node[main] [below right of=5] (9) {$C_m$}; 
            \draw[->] (4) -- node[midway, above, sloped, pos=0.5] {$\text{\small type}$} (6);
            \draw[->] (4) -- node[midway, below, sloped, pos=0.5] {$\text{\small type}$} (7);
            \draw[->] (5) -- node[midway, above, sloped, pos=0.5] {$\text{\small type}$} (8);
            \draw[->] (5) -- node[midway, below, sloped, pos=0.5] {$\text{\small type}$} (9);
            
            \node at ($(2)!.45!(3)$) {\vdots};
            \node at ($(4)!.45!(5)$) {\vdots};
            \node at ($(6)!.52!(7)$) {\ldots};
            \node at ($(8)!.52!(9)$) {\ldots};
        \end{tikzpicture} 
        \caption{SchemEX \mschemex{}}
        \label{fig:collections_sub4}
    \end{subfigure}
    \caption{The summary models \mattrcol{}, \mclasscol{}, \mpropertytypecol{}, and \mschemex{} considered in this work. Here, $x$ is the root vertex that is to be summarized, $y$ is an object connected to $x$ through an attribute set, and $C$ is an object connected to $x$ or $y$ connected via an \texttt{rdf:type} edge.}
    \label{fig:collections}
\end{figure}

For the graph summary model \AttributeCollection{} \mattrcol{} (see \autoref{fig:collections_sub1}), the information used as described above is the property set. 
The property set of a vertex is its set of outgoing edge labels, excluding \texttt{rdf:type}. 
The \ClassCollection{} \mclasscol{} (see \autoref{fig:collections_sub2}) can be seen as the other side of the \AttributeCollection{}. The information used is called the type set and consists only of the RDF types. 
A combination of both is the \PropertyTypeCollection{} \mpropertytypecol{} shown in \autoref{fig:collections_sub3}. 
With the gathered knowledge, \Schemex{} \mschemex{} can be described as a combination of the specifications of multiple summary models. It is shown in \autoref{fig:collections_sub4}.
For \mschemex{}, the equivalence class is calculated by the aggregation of the \mclasscol{} information on the root vertex and the \mclasscol{} information of a neighborhood. The neighborhood is defined by the \mattrcol{} specification of the root vertex.
Thus, we are aggregating the \mclasscol{} information of the root vertex and the \mclasscol{} information of the neighboring vertices into a single list. The calculation of the hash for the equivalence class is then the same as for the other models.

\subsection{Graph Neural Networks Selected for the Experiments}
\label{subsec:gnnsM}
We describe the graph neural networks (GNNs) we selected from \autoref{subsec:gnnRW} that we use in our experiments.
These are GCN, GraphSAGE, GraphSAINT, and \GraphMLP.
For all GNNs, we use a ReLU-activation function and dropout for all our hidden layers, while a softmax function is used on the output layer. 
One common issue with GNN approaches is that they are highly sensitive to the vertex degrees, which can lead to numerical instabilities and problems during training~\cite[Chapter~5.2]{hamilton2020graph}. This can usually be solved by normalizing w.r.t.\@ degree prior to aggregation. 
However, we cannot do this, as the goal of graph summaries is to preserve the graph structure and the vertex degree is an important structural feature.

The optimization objective during training is a negative log likelihood loss function.
Given the high memory requirements of GNNs, we apply a sampler on our extensive dataset.
Here, we are following GraphSAINT's subgraph sampling strategy. 
We use a directed-edge sampler to reduce the batch size and then apply a GNN on a batch of root-vertex centered subgraphs. 
This is done via semi-supervised transductive training~\cite[Chapter~6.1.1]{hamilton2020graph}: during inference all vertices are present, while the loss function is only calculated on the labeled vertices, which are the root vertices.

\extended{We report the results per GNN model as \textit{GNN}-$k$ with \textit{GNN} corresponding to the neural model and $k$ to the number of hops.}

\paragraph{GCN} 
\label{par:GCNM}
To apply classic GCNs, we chose to create the models on the full adjacency matrix but only feed batches created by our sampler through the models. 
We do not shrink the adjacency matrix to the batch size, because of the high computational complexity of reintegrating those smaller graphs into the complete model. 
\paragraph{GraphSAGE}
In our application, we use a vanilla GraphSAGE with a mean-aggregator. 
One limitation of GraphSAGE is that it assumes that nodes in a neighborhood belong to the same equivalence class, which is not guaranteed in our datasets. But as different applications~\cite{hamilton2018inductive} show, this limitation can be ignored.

\paragraph{GraphSAINT}
We replace GraphSAINT's vertex sampler by one based on the inverse class distribution. 
\shortorextended{We apply a GraphSAINT network with 2 layers of Weisfeiler–Leman kernels with jumping knowledge.}{We apply a GraphSAINT network using Weisfeiler–Leman kernels. If the network has two layers then we also apply jumping knowledge.}

\paragraph{\GraphMLP}
We use \GraphMLP~\cite{graphMLP} and remove the normalization layer as above.
\extended{The normalizing technique, applied by the authors, is again skipped because of the characteristics of our application.}
For our experiments we align the $r$-hop-parameter of \GraphMLP to the $k$-hop characteristic of each graph summary model. 
The specific $r$-value is appended to the model name.

\paragraph{Discussion of Further Models}
We also considered using R-GCN and SGC.
But both models could not be used on the large numbers of classes of our datasets.
Our DyLDO dataset has about $7$ times more vertices, $20$ times more edges, $110$ times more edge types, and $10,000$ times more classes than the datasets used in the original R-GCN paper by Schlichtkrull et al.~\cite{schlichtkrull2017modeling}.
Since the complete R-GCN model must be present during training and R-GCN has additional weights per edge type, the model massively exceeds the available GPU memory, even with weight sharing techniques such as basis decomposition~\cite{schlichtkrull2017modeling}.
We ran pre-experiments for SGC on a smaller dataset.
SGC scored considerably lower than the other GNNs and was also the biggest model, which pushed us to the limits of our GPU memory. 
In our main experiments on the larger DyLDO dataset, the SGC model exceeded the GPU memory. 
Thus, we do no use R-GCN and SGC\extended{~in our experiments}. 
\extended{For more details see \autoref{sec:appendixSGCN}.}

\subsection{Baselines}
\label{subsec:baselineM}

We consider a simple but effective neural baseline and a strong non-neural baseline.

\paragraph{MLP}
\label{subsubsec:mlpM}
We use a standard multi-layer perceptron (MLP) with two hidden layers and ReLU-activation function with dropout. This is motivated by the strong performance of MLPs for text classification~\cite{galke2021forget}.

\paragraph{Bloom Filter}

As a non-neural baseline, we use a Bloom filter. For each vertex, the graph summary specific information is added into an empty Bloom filter array. To calculate the equivalence class of the vertex, the Bloom filter array is hashed. This approach avoids the sorting normally required for graph summarization. However, it comes with a cost in terms of accuracy as the Bloom filter is susceptible to false positives, meaning that some originally different equivalence classes result in the same Bloom filter array and thereby are wrongly assigned to the same equivalence class. 

We define our Bloom filter with $4$, $15$, and $60$ input items, corresponding to the $75$th, $95$th, and $99$th percentile of the node degree distribution of the DyLDO dataset and false positive probabilities of 
$10^{-3}$ and $10^{-1}$. 
\extended{These combinations result in different numbers of hash functions and bits in the array. 
Results of more configurations are documented in Appendix~\ref{sec:addStatsAppendixBloom}.}

\section{Experimental Apparatus}
\label{sec:expApparatus}

\extended{We introduce the dataset, hyperparameter optimization, the experimental procedure, important information regarding the implementation, and the evaluation metrics.}

\subsection{Datasets}
\label{subsec:datasetEA}

\extended{This section gives an overview over the considered RDF-dataset  and the class distributions generated by the different graph summary models.}

\label{subsubsec:datasetStatisticsEA}
\subsubsection{Basic Statistics}
The Dynamic Linked Data Observatory (DyLDO)~\cite{kafer2012towards} is a framework for monitoring linked data. 
It takes snapshots of a subset of linked data, to capture its dynamics. 
\extended{DyLDO takes a sampling-based approach of seed-URIs, which are also representative for the vertices of the graph.}
Each weekly snapshot has about $100$ million triples.
The dataset is populated by crawling from a list of seed-URIs. 
Crawling from the seed URIs is restricted to a depth of two hops~\cite{kafer2012towards}. 
We use this dataset because of its reasonable size, but it still reflects real-world linked data. 

\paragraph{DyLDO}
The main snapshot used for this paper is the very first one, as it is with $127M$ triples the largest one.
Analysis by Blume and Scherp~\cite{blume2020indexing} shows that it contains $7{\small,}093{\small,}011$ vertices with $15{\small,}017$ overall edge properties. From these edge properties, \texttt{rdf:type} occurs $5.4\,\mathrm{M}$ times. Each vertex has on average $17$ outgoing properties.

\paragraph{\dyldoschemex{}}
The SchemEX graph summary model \mschemex{} is the only one we consider that requires a second hop, so there is a huge increase in observed vertices and consequently in the size of each subgraph. Hence, generating and using all $2$-hop subgraphs during inference would have exceeded our disk space and memory capacity.
Therefore, a smaller portion of the dataset was used to reduce the computational requirements. 
This smaller dataset was built by removing  vertices with a class occurrence of $<6$, which leaves around $15\%$ of the dataset\extended{~and fits our hard drive restrictions}.
Also only subgraphs with fewer than $500$ vertices are kept.
In our sampling process (during inference),  we use this number as guard condition for the maximum size of the mini-batch (see Section~\ref{sec:generate-mini-batches}).
We call this dataset \dyldoschemex{} and likewise we call \mschemex{} on this dataset \mschemexused{}. 
This dataset has $16.7\,\mathrm{M}$ triples, $5{\small,}312{\small,}991$ vertices, and $8{\small,}102$ edge properties. 
From these edge properties, \texttt{rdf:type} occurs $3.7M$ times. Each vertex has on average $3$ outgoing properties.

\subsubsection{Class Distributions}
\label{subsubsec:datasetClassDistributionEA}
Computing the summaries for the four summary models introduced in \autoref{subsec:graphSummaryM}, generates four different datasets on which to train the neural networks. 
These four graph datasets differ in the number of equivalence classes, which correspond to the number of classes for the graph neural networks.
\autoref{tab:numberEquiClasses} shows the number of classes for each summary model.

\begin{table}[htb]
\small
    \centering
    \begin{tabular}{
    |r|r|r|r|r|}
        \hline
         \multicolumn{1}{|l|}{\mattrcol{}}   &\multicolumn{1}{l|}{ \mclasscol{}}  &\multicolumn{1}{l|}{ \mpropertytypecol{}}   &\multicolumn{1}{l|}{ \mschemex{} }  &\multicolumn{1}{l|}{ \mschemexused{} }  \\
        \hline
        $162{\small,}521$     & $576$         & $178{\small,}472$             & $335{\small,}608$     & $18{\small,}314$         \\
        \hline
    \end{tabular}
    \caption{Number of equivalence classes in the DyLDO dataset for each summary model. \mschemexused{} denotes the number of \mschemex{} classes in the \dyldoschemex{} dataset.}
    \label{tab:numberEquiClasses}
\end{table}

\begin{figure}[htb]
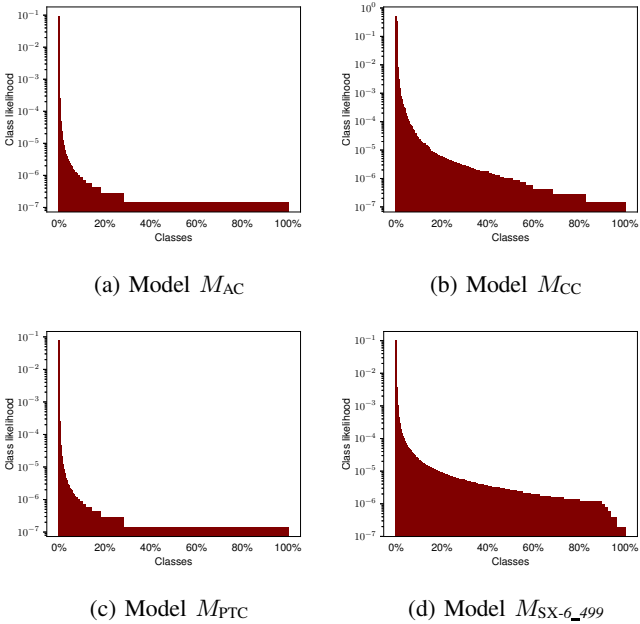

    \centering
    \begin{subfigure}{.24\textwidth}
        \begin{center}
            \resizebox{1.\columnwidth}{!}{\input{ressources/class_distribution_01.pgf}}
        \end{center}
        \caption{Model \mattrcol{}}
        \label{fig:class_distr_mac}
    \end{subfigure} 
    \begin{subfigure}{.24\textwidth}
        \begin{center}
            \resizebox{1.\columnwidth}{!}{\input{ressources/class_distribution_02.pgf}}
        \end{center}
        \caption{Model \mclasscol{}}
        \label{fig:class_distr_mcc}
    \end{subfigure}
    \begin{subfigure}{.24\textwidth}
        \begin{center}
            \resizebox{1.\columnwidth}{!}{\input{ressources/class_distribution_04.pgf}}
        \end{center}
        \caption{Model \mpropertytypecol{}}
        \label{fig:class_distr_mptc}
    \end{subfigure} 
    \begin{subfigure}{.24\textwidth}
        \begin{center}
            \resizebox{1.\columnwidth}{!}{\input{ressources/class_distribution_03_min_support_5_1_500.pgf}}
        \end{center}
        \caption{Model \mschemexused{}}
        \label{fig:class_distr_ms}
    \end{subfigure} 
\caption{The class distributions of the different graph summary models on the DyLDO dataset. The horizontal axis shows the classes sorted in descending likelihood.}
\label{fig:ClassDistributions}
\end{figure}

In \autoref{fig:ClassDistributions}, we plot the class distribution for each summary model as the likelihood of a specific class appearing.
The class distributions of \mattrcol{} (\autoref{fig:class_distr_mac}), \mclasscol{} (\autoref{fig:class_distr_mcc}), \mpropertytypecol{} (\autoref{fig:class_distr_mptc}), and \mschemexused{} (\autoref{fig:class_distr_ms}) all show a skewed distribution due to the majority of classes appearing only once. 
\extended{For reference, the class distribution of  the full \mschemex{} can be seen in Appendix~\ref{sec:addStatsAppendixMsWhole}.}

\subsection{Hyperparameter Optimization}
\label{subsec:hyperparameter}

We applied a hyperparameter search to the learning rate over the values in $\{0.1, 0.01, 0.001\}$ of the Adam optimizer, dropout in $\{0.0,0.2,0.5\}$, and the hidden layer size $\{32,64\}$ for the GNNs with $2$-hops. For \GraphMLP, we apply the hyperparameter search onto the weighting coefficient $\alpha\ \in \{1.0,10.0,100.0\}$,  the temperature $\tau\ \in \{0.5,1.0,2.0\}$, learning rate in $\{0.1, 0.01\}$, and the hidden layer size $\{64,256\}$. The validation accuracy and training loss are monitored and evaluated on the details of convergence speed and stability for this purpose. 
For the search, we trained our models for $30$ epochs.
We manually confirmed the optimization process by inspecting the loss curve. 
This ensured that the models properly converged. 
The validation accuracy is the mean over $15$ mini-batches from the validation fold. 

For our MLP, we use a hidden layer size of $1{\small,}024$ following prior work~\cite{galke2021forget}.
For \GraphMLP, we use the 
recommended dropout of $0.6$\extended{~without any hyperparameter optimization of the dropout value}.
For the GNNs, the dropout values are overall low.
Interestingly, some of our GNN models work best without any dropout.
\shortorextended{The optimal hyperparameter values and parameter counts for the competing models are given in our extended report~\cite{DBLP:journals/corr/abs-2203-05919}.}{The optimal hyperparameter values of the GNNs for the different summary models are shown in Appendix~\ref{sec:addStatsAppendixHyper}.
An overview over all parameter counts for the competing models is given in Appendix~\ref{sec:parameter-count-main-experiments}.}

\subsection{Procedure}
\label{subsec:procedure}
Our goal is to use the equivalence classes of the graph summaries as labels for our classification task. 
Given our large dataset, we had to apply a sampling method and train our models on mini-batches. 
To apply the GNNs on our data, we have to use the following preprocessing steps. Afterwards, we explain the training and the evaluation process.

\paragraph{Preprocessing}
In the first step, all triples $(s,p,o)$ in $G$ are preprocessed into a dictionary. This dictionary consists of the subject URI as key and the corresponding information structure for each subject vertex as value. 
The latter contains the equivalence class label, all the neighbors, and the fold number of a subject vertex. 
Based on this dictionary, the chosen graph summary model is calculated. The calculated hash values (equivalence classes) are transformed into class labels $\vy$, which annotate the corresponding vertices\extended{~and are set in \textit{SubjectInformation}}. 
To ensure the separation of training, validation, and test data, we split our data into ten folds by assigning each subject vertex randomly to a fold. So a fold consists of the subgraphs of its subjects. 
\extended{
The pseudocode of the algorithm to generate the training data for the different summary models is shown in Appendix~\ref{sec:GeneratingTrainingDataForTheSummaryModels}.}

The structure for each subgraph consists of a feature matrix, a list of class labels for each node, an adjacency list, and a list containing the edge types, which corresponds to the adjacency list element. 
The feature matrix one-hot encodes each node in the subgraph. Because of RAM limitations on the GPU, we apply a sampler to generate mini-batches (see below).

\paragraph{Training the Neural Networks}
After the preprocessing, the neural networks are trained on mini-batches from the training folds and then tested on the mini-batches from the test fold. The training process is monitored and evaluated on the validation accuracy.

\begin{algorithm}[htb]
\small
	\caption{Create mini-batch for training from the foldset} 
	\label{alg:CreateMiniBatches}
	\begin{algorithmic}[1]
	\Procedure{CreateMiniBatch}{$f$: foldset, $guard$: integer}
    	\State GraphData $batch \leftarrow \emptyset$
	    \State Draw subgraph $sample$ from $f$
    	\While{ $batch.\#nodes + sample.\#nodes < guard$ }
	        \State Append $sample$ to $batch$
	    \State Draw next subgraph $sample$ from $f$
	    \EndWhile
        \State \Return $data$
    \EndProcedure
	\end{algorithmic} 
\end{algorithm}

\label{sec:generate-mini-batches}

Our sampler differs from the vertex-, edge-, and random walk-based samplers used in GraphSAINT~\cite{DBLP:journals/corr/abs-1907-04931}.
Independent of the specific sampling method, GraphSAINT determines the vertices involved by the sample (\eg the edges' source and target vertices) and adds the subgraphs induced by those vertices to the batch.
This can cause a huge variance in the number of vertices and edges, \ie huge differences in the batch sizes.
In practice, this lead to out of memory problems on the GPU, so we restrict the size of the batch with the help of a guard.

To generate the mini-batches with a constant mini-batch size $guard$, we apply \autoref{alg:CreateMiniBatches} with a set of folds $f$ and the $guard$ condition representing the desired batch size. 
To maintain a constant mini-batch size, the mini-batches are padded with additional \emph{dummy} vertices that are single nodes without any edges and a class label of $-1$. 
This is done until reaching the $guard$ limit. 
We define $guard = 500$. 
This restriction of the constant mini-batch size is based on the limitation of the GPU memory. 
The goal is to fit the largest possible mini-batch into memory for all GNNs. 
As an optimizer Adam is applied for all GNNs. 

Only the root vertices of the subgraphs are considered in the loss and accuracy calculation, ignoring the object neighbors and the dummy vertices. To compensate for the skewed class distribution, the subgraphs are randomly sampled for the training process by a probability which is inversely proportional to the frequency of the class occurrences. This results in a uniform distribution of sampled subgraphs to increase stability and speed of training. In the last step, those mini-batches can be used in the training process or in validation or test of the model.
\extended{For the training of the models, the results of the hyperparameter optimization, \autoref{subsec:hyperparameter}, were used.}
The models are trained for $30$ epochs, except for \GraphMLP-1 and \mclasscol{}, which were trained for $25$ epochs.

\paragraph{Evaluation}
To evaluate our methods, we run a full $10$-fold cross-validation. For each step, we trained our GNN models on the data of $8$ folds, evaluated the training progress on $1$ validation fold, and tested it on $75$ mini-batches of $1$ test fold to report a mean testing accuracy per run. 

\subsection{Metrics}
\label{subsec:metrics}

We use accuracy to measure the classification performance of the GNNs. We report the mean and standard error of the test accuracy on a 10-fold cross-validation. 

The Bloom filter is evaluated based on accuracy and impurity. 
We compare between the hash values calculated via the graph summarization method from \autoref{subsec:graphSummaryM} as ground truth and the hash values calculated by the Bloom filter method (see \autoref{subsec:baselineM}) of all $N$ subject vertices, which we call the root set $\gR$. 
We cluster the vertices in $\gR$ via their Bloom filter hash values, resulting in the clustering $\Omega$. 
Each cluster $\omega \in \Omega$ can be clustered again based on the ground truth values of each subject vertex, resulting in the clusters $\gC$. We compute the accuracy based on the assumption that the vertices in the largest category per cluster are the true positives. This follows the fundamentals of the impurity definition.
\begin{comment}
For the Rand agreement-ratio $I_{Rand}(\Omega)$ we consider pairs of vertices $v_i,v_j$ and their respective Bloom filter values $b_i,b_j$ and ground truth values $g_i,g_j$. These are compared based on the number of agreements and disagreements resulting in \autoref{eqn:Rand}. We count a pair as an agreement when $(b_i = b_j \wedge g_i = g_j) \vee (b_i \neq b_j \wedge g_i \neq g_j)$ holds. If this does not hold we count it as disagreement.
\end{comment}
Per cluster $\omega \in \Omega$, the impurity measure $Q$ is calculated based on the Gini-index resulting in $Q_g$~\cite{Gini} and the probability $p_{ij} = \frac{m_{ij}}{|\omega_i|}$ with $m_{ij}$, the number of objects from $C_j$ in $\omega_i$. 
The final impurity value $Q_g(\gR)$ 
is calculated via the sum over all the clusters weighted by their relative frequency $\frac{|\omega_i|}{N}$ with
    \begin{equation*}
        Q_g(\gR) = \sum_{i=1}^{|\Omega|} \frac{|\omega_i|}{N} Q_g(\omega_i)
        \label{eqn:Gini}
\text{,} \quad
        Q_g(\omega_i) = 1 - \sum_{j=1}^{|\gC|} p_{ij}^2 \,.
    \end{equation*} 

\section{Results}
\label{sec:results}

For our results, we applied $10$-fold cross-validation for all our models except the Bloom filter, and all summary models. We did not cross-validate Bloom filters, as they are deterministic and would give the same result each time. The results can be seen in \autoref{tab:results}.

\begin{table*}[htb]
\newcommand{\pad}{\phantom{{}\pm 0.0000}}
\newcommand{\nodata}{$\phantom{0.0000}-\phantom{0.0000}$}
    \centering
    \begin{tabular}{|l||r|r|r|r|}
        \hline
         \multicolumn{1}{|l||} {}  & \multicolumn{1}{l|} {\mattrcol{} }  &\multicolumn{1}{l|} { \mclasscol{} } & \multicolumn{1}{l|} {\mpropertytypecol{} }  & \multicolumn{1}{l|} {\mschemexused{}}   \\
        \hline
        Bloom filter ($n=4$, $p=10^{-1}$)   & $0.8376\pad$                      & $\topOne{0.8562}\pad$             & $0.7987\pad$                      & $0.7598\pad$                      \\
        Bloom filter ($n=15$, $p=10^{-1}$)  & $0.8555\pad$                      & $0.8562\pad$                      & $0.8216\pad$                      & $0.7764\pad$                      \\
        Bloom filter ($n=15$, $p=10^{-3}$)  & $0.8562\pad$                      & $0.8562\pad$                      & $0.8223\pad$                      & $\topOne{0.7767}\pad$             \\
        Bloom filter ($n=60$, $p=10^{-1}$)  & $0.8562\pad$                      & $0.8562\pad$                      & $\topOne{0.8224}\pad$             & $0.7767\pad$                      \\
        Bloom filter ($n=60$, $p=10^{-3}$)  & $\topOne{0.8563}\pad$             & $0.8562\pad$                      & $0.8224\pad$                      & $0.7767\pad$                      \\
        \hline
        MLP-2                               & $\topFour{0.6033} \pm 0.0048$ & $0.8496 \pm 0.0067$           & $\topThree{0.5706} \pm 0.0030$& $\topFour{0.5470} \pm 0.0127$ \\
        \hline
        GCN-2                  & $0.6007 \pm 0.0060$           & $\topTwo{0.8559} \pm 0.0025$  & $0.5611 \pm 0.0047$           & $\topThree{0.5473} \pm 0.0052$\\
        GraphSAINT-2                        & $0.5858 \pm 0.0047$           & $\topFour{0.8538} \pm 0.0026$ & $0.5447 \pm 0.0062$           & $0.5080 \pm 0.0076$           \\
        GraphSAGE-2                         & $\topThree{0.6138} \pm 0.0052$& $\topThree{0.8541} \pm 0.0027$& $\topFour{0.5652} \pm 0.0079$ & $0.5354 \pm 0.0053$           \\
        \hline
        \GraphMLP-1                          & $\topTwo{0.6230} \pm 0.0057$  & $0.8401 \pm 0.0031$           & $\topTwo{0.5968} \pm 0.0037$  & \nodata                           \\
        \GraphMLP-2                          & \nodata                          & \nodata                           & \nodata                           & $\topTwo{0.5489} \pm 0.0066$  \\
        \hline
    \end{tabular}
    \caption{Results for $10$-fold cross-validation on DyLDO's first snapshot reported by mean accuracy with standard error (higher accuracy, lower standard error better). The chosen $n$-values for Bloom filter describe the $75$th, $95$th, and $99$th percentile of the node degree distribution. The top four models are highlighted per graph summary: \topOne{first}, \topTwo{second}, \topThree{third}, and \topFour{fourth} place. For Bloom filter the model with the highest accuracy score, with the smallest number of hash functions $k$, and the smallest number of bits in the array $m$ is marked as best. Best viewed in color.}
    \label{tab:results}
\end{table*}

The Bloom filter outperforms all tested neural models, by a margin that depends on the summary model.
The second-best result for \mattrcol{} is $0.2333$ lower (\GraphMLP-1), \mpropertytypecol{} is $0.2256$ lower (\GraphMLP-1) and \mschemexused{} is $0.2278$ lower (\GraphMLP-2). For \mclasscol{} the Bloom filter and sample-based GCN-2 (short: GCN-2) performance is similar.
The results for the neural network models of \mattrcol{} are within $0.0372$, \mclasscol{} is within $0.0158$, \mpropertytypecol{} is within $0.0521$, and \mschemexused{} is within $0.0409$ of the reported accuracy. The standard error for all GNN models is below $0.0080$, with the exception of MLP-2 and \mschemexused{}. The lowest standard error is $0.0025$ for sample-based GCN-2 and \mclasscol{}. 

\begin{table}[htb]
    \centering
    \begin{tabular}{|l||r|r|r|r|}
        \hline
         \multicolumn{1}{|l||} {}  & \multicolumn{1}{l|} {\mattrcol{} }  &\multicolumn{1}{l|} { \mclasscol{} } & \multicolumn{1}{l|} {\mpropertytypecol{} }  & \multicolumn{1}{l|} {\mschemexused{}}   \\
        \hline
        $n=4$, $p=10^{-1}$      & $0.2232$      & $0.2423$      & $0.2793$              & $0.2363$          \\
        $n=15$, $p=10^{-1}$     & $0.2026$      & $0.2423$      & $0.2526$              & $0.2202$          \\
        $n=60$, $p=10^{-1}$     & $0.2018$      & $0.2423$      & $0.2517$              & $0.2198$          \\
        \hline
    \end{tabular}
    \caption{Impurity metric for Bloom filter using the Gini-index with expected input items $n$ and false positive probability $p$ (lower impurity better).}
    \label{tab:BloomImp}
\end{table}

\autoref{tab:BloomImp} shows the impurity metric for the differently parameterized Bloom filters.
The Bloom filter accuracy results for every parameter combination within a single summary model differ at most by $0.0237$ from each other (see \autoref{tab:results}).
For the impurity measures, a similar result can be observed. The average scores are very close to each other. The higher the chosen $n$ and the lower the chosen $p$, the better the accuracy and impurity results. All Bloom filter accuracies and impurities for \mclasscol{} are equivalent. 

\section{Discussion}
\label{sec:discussion}

\paragraph{Main Results}
Our experiments show that graph summarization via vertex classification is a challenging task.
The models have to deal from a few hundred to hundreds of thousands of classes, depending on the summary model.
Bloom filters deliver overall the best and most consistent results. 
The graph neural networks are on par with Bloom filter in one of the four summary models, namely \mclasscol.
The number of classes for the \mclasscol{} model is rather low ($576$). 
Generally, the performance of the graph neural networks are close to each other.
For each summary model, the difference between the best- and worst-performing GNN is less than $4.1$ points.

Our results show that \GraphMLP outperforms the other neural models in three out of four graph summary experiments. 
However, it is difficult to make a decisive conclusion as to which graph neural network model is the best, as Shchur et al.~\cite{pitfalls} have shown that minor changes in the model parameters, training data, etc.\@ of GNNs have a significant effect on the ranking of the models.
The \GraphMLP model, which does not use message passing and introduces an MLP-like structure, was only tested on smaller datasets in the original paper~\cite{graphMLP}. 
Thus, for the first time, it has been applied on large graphs with large numbers of classes.
Furthermore, GraphSAGE outperforms GraphSAINT in all cases. 
The good performance of GraphSAGE (among the top-3 neural network models in three out of four graph summary experiments) on a classification task where the labels are dependent on the neighborhood structure might be surprising, but is consistent with findings in previous works~\cite{pitfalls}. 
The random neighborhood sampling and the learning of the aggregation function of GraphSAGE seem to be strong regularizers.

Our baseline MLP is quite strong but does not outperform graph-based models. 
Such a strong performance of a standard MLP may be surprising, but MLP has performed well in other classification tasks~\cite{DBLP:conf/ijcnn/VenugopalTFS21,galke2021forget}.

\paragraph{Further Discussion}

The different summary models have different numbers of class labels (see \autoref{tab:numberEquiClasses}). 
This means that with an increasing number of equivalence classes, the complexity of classification also increases because it requires more differentiation. 
The influence of complexity can be observed in two ways. 
The accuracy results of our different neural network models are worse for models with a higher class count and these results differ more within models with a higher class count.
This can be best observed in the good and very similar results for \mclasscol{}, as \mclasscol{} has the lowest number of classes by a large margin (see \autoref{tab:numberEquiClasses}).
An exception to this observation is the results for \mschemexused{}. We suspect that this deviation is caused by the $2$-hop property of \Schemex{} in conjunction with the lack of normalization. The $2$-hop characteristic increases the complexity significantly by aggregating additional information from the second hop into the root node.

The choice of parameters for the Bloom filter does not seem to have much impact. Furthermore there seem to be diminishing returns when higher $n$ and the lower $p$ are chosen. The Bloom filter accuracy results seem to converge at ${\sim}0.8563$ for \mattrcol{}, 
${\sim}0.8562$ for \mclasscol{}, ${\sim}0.8224$ for \mpropertytypecol{}, and ${\sim}0.7767$ for \mschemexused{}. The impurity results converge at ${\sim}0.2018$ for \mattrcol{}, ${\sim}0.2423$ for \mclasscol{}, ${\sim}0.2517$ for \mpropertytypecol{}, and ${\sim}0.2198$ for \mschemexused{}. 
This strong performance can be explained because Bloom filters approximate set membership with an one-sided error, which is required by various summary models. The high number of bits and the resulting high number of potential representation of classes can also contribute the the strong performance.

\paragraph{Threats to Validity and Future Work}

It is known that experiments on GNNs can suffer from parameter choices~\cite{pitfalls}.
We address these challenges by running a $10$-fold cross validation and provide the standard errors of our GNN results.
As shown in \autoref{tab:results}, the standard error over our folds is very low and consistent within each summary model.
Another observation is that the Bloom filter scores for \mclasscol{} are identical. 
This may have suggested some error in our procedure. 
However, we manually checked the results. 
Furthermore, we executed the same experimental procedure (code, evaluation etc) for \mclasscol{} as for the other summary models, which further supports the correctness of the \mclasscol{} results.

One may note that the applied GNN models consider directed edges as undirected, which is commonly the case in this domain. 
This limitation might have impacted our classification performance, because graph summaries are sensitive to the edge direction. 
However, as discussed in \autoref{subsec:gnnsM} our graph sampler applied to feed the GNNs considers the direction of the edges based on the applied summary models.
Thus, the applied GNNs also properly consider the edge directions when computing the classifications of the vertices.

We experiment with real-world datasets. 
Thus, the number of classes is much higher than in commonly used datasets~\cite{DBLP:journals/corr/YangCS16}, which are based on citations. 
Approximately $54\%$ of our classes only occur once in the dataset. This leads to an extremely skewed class distribution, as can be seen in \autoref{fig:ClassDistributions}.
Thus, overall one can consider the results on such skewed datasets as impressive and further improvements could be made by, \eg using hybrid models of GNNs combined with label propagation~\cite{WangLeskovec2022}.

\extended{With our datasets we already consider graphs that are orders of magnitude larger in terms of classes and relations (see Section~\ref{sec:expApparatus}).
The biggest dataset for R-GCN~\cite{schlichtkrull2017modeling} has $1.67M$ vertices, with $1,000$ vertices being labeled and having $11$~classes and $133$ relations. 
While trying to apply the R-GCN model to our datasets, we ran into memory limitations.
In the future work, we like to experiment with further datasets such as citation graphs~\cite{DBLP:journals/corr/YangCS16}.}

\extended{To gain further insight into the performance and applicability of GNNs for graph summarization on larger datasets, our experiments should be repeated on further datasets, like the Billion Triple Challenge~\cite{herrera2019btc} or the LOD Laundromat~\cite{beek2014lod}.}

\extended{A huge performance improvement would be to follow the proposed method of GraphSAINT, where only the partial model with the required nodes of the mini-batch are present. This would reduce the model's memory footprint and would allow a higher mini-batch size. For this, a method is required which manages the parameter transfer from the complete model, tracking the whole process, to the partial model, on which inference and back-propagation can be run, and vice versa.
This improvement would allow to apply models, like R-GCN and SGC.}

\extended{We ran our experiments without normalization, because the class labels are dependent on the structural information of the neighborhood. Normalization and regularization could also be the subject of further research. This is motivated by the deviating results we received for the \mschemexused{}.}
\extended{
Another future study could be a runtime analysis in which
we perform a cost-benefit analysis of using GNN methods with the involved the accuracy loss compared to the computation performance of lossless computations.}

\section{Ablation Study: No Singleton Classes}
\label{subsec:ASNoSingletons}

\extended{We provide further insights into our experimental results by analyzing the impact of removing the singleton classes from the graphs on the models' performance.
We run additional ablations and analyses, which are reported in Appendix~\ref{sec:AblationStudy}.}

To check the effect of the skewed class distribution on the classification accuracy, we run an ablation study removing all singleton class occurrences from our graphs (min-support $> 1$). We call the models with removed singletons \mattrcolns{}, \mclasscolns{}, and \mpropertytypecolns{}. 
On average, $46\%$ of the classes remain (see \autoref{tab:numberEquiClassesSing}). 
\extended{In Appendix~\ref{sec:addStatsAppendixNSClassDistr}, the new class distribution can be seen, in comparison to the original class distribution.}
Because the data distribution changes in this procedure, another hyperparameter optimization is applied.
We did not include \Schemex{}, because in the dataset used for \mschemexused{}, \dyldoschemex{}, we are already filtering with a min-support higher than $1$.
\shortorextended{The optimal hyperparameter values and parameter counts for these models without singletons can be found in our extended report~\cite{DBLP:journals/corr/abs-2203-05919}.}{
The optimal hyperparameter values can be seen in Appendix~\ref{sec:addStatsAppendixNSHyper}.
The parameter counts for these models without singletons can be found in Appendix~\ref{sec:addStatsAppendixNSParamCount}.}

\begin{table}[htb]
    \centering
    \begin{tabular}{|l||r|r|r|}
        \hline
         \multicolumn{1}{|l||} {}  & \multicolumn{1}{l|} {\mattrcolns{} }  &\multicolumn{1}{l|} { \mclasscolns{} } & \multicolumn{1}{l|} {\mpropertytypecolns{} } \\
        \hline
        Remaining Classes   & $45{\small,}198$      & $474$         & $49{\small,}719$              \\
        \hline
        Class Percentage    & ${\sim}28\%$   & ${\sim}82\%$   & ${\sim}28\%$           \\
        \hline
        Remaining Subgraphs & $6{\small,}975{\small,}688$   & $7{\small,}092{\small,}909$   & $6{\small,}807{\small,}122$           \\
        \hline
        Subgraph Percentage & ${\sim}98\%$   & ${\sim}100\%$  & ${\sim}96\%$           \\
        \hline
    \end{tabular}
    \caption{Number of equivalence classes in the DyLDO dataset by summary model, after the removal of classes that only occur once, the percentage of remaining classes (cf. \autoref{tab:numberEquiClasses}), the number of the original $7,093,011$ subgraphs remaining, and the percentage of the remaining subgraphs.}
    \label{tab:numberEquiClassesSing}
\end{table}

\begin{table*}[htb]
    \centering
    \begin{tabular}{|l||r|r|r|}
        \hline
         \multicolumn{1}{|l||} {}  & \multicolumn{1}{l|} {\mattrcolns{} }  &\multicolumn{1}{l|} { \mclasscolns{} } & \multicolumn{1}{l|} {\mpropertytypecolns{} } \\
        \hline
        MLP-2               & $\topTwo{0.6501} \pm 0.0059$  & $\topTwo{0.8298} \pm 0.0022$  & $0.6050 \pm 0.0078$           \\
        \hline
        GCN-2  & $0.6298 \pm 0.0031$           & $0.8129 \pm 0.0069$           & $\topTwo{0.6154} \pm 0.0029$  \\
        GraphSAINT-2        & $0.6426 \pm 0.0038$           & $\topThree{0.8234} \pm 0.0027$& $0.5952 \pm 0.0067$           \\
        GraphSAGE-2         & $\topThree{0.6442} \pm 0.0039$& $\topOne{0.8300} \pm 0.0022$  & $\topThree{0.6141} \pm 0.0022$\\
        \hline
        \GraphMLP-1          & $\topOne{0.6627} \pm 0.0039$  & $0.8224 \pm 0.0024$           & $\topOne{0.6213} \pm 0.0042$  \\
        \hline
    \end{tabular}
    \caption{Results for $10$-fold cross-validation for the different models after the removal of singleton classes reported by mean accuracy and standard error (higher accuracy, lower standard error better). The top three models are highlighted per graph summary: \topOne{first}, \topTwo{second} and \topThree{third} place. }
    \label{tab:resultsNoSingletons}
\end{table*}

The results are reported in  \autoref{tab:resultsNoSingletons} via the average and standard error of the test accuracy for a $10$-fold cross-validation after training for $40$ epochs.
There is an increase in accuracy for \mattrcolns{} and \mpropertytypecolns{} by ${\sim}3$--$6\%$. Also the standard error for \mattrcolns{} is lower except for MLP-2.
However, \mclasscolns{} gets worse by ${\sim}2$--$4\%$. 
The singleton class occurrences make the problem harder. Comparing the results of this ablation study and the experiment, the difference is actually smaller than expected.
From the results it is noticeable that the models with the lowest hidden layer size also have the lowest accuracy: for \mattrcolns{}, it is GCN-2 and GraphSAINT-2 for \mpropertytypecolns{}. This was also a finding in the main experiment.
Even without the singletons, we still have a very skewed class distribution.
Since we have a train--validate--test split on the data, further studies could be done on a dataset with min-support higher than $2$ to consider the random distribution of classes in the data split. 

\extended{Also, the lower performance of GNNs on the dataset without singletons for the \mclasscolns{} should be researched as it is the only result that contradicted our expectation. 
To combat the skewed class distribution another less invasive measure could also be developed, which does not remove single class occurrence nodes but which maps the class onto the next best-fitting class~\cite{DBLP:conf/kcap/GottronSKP13}.}

\section{Conclusion}
\label{sec:conclusion}

We methodically compared the effectiveness of computing graph summaries, using established and recent graph neural network models and Bloom filters.
Our main result is that Bloom filters outperformed the GNN models by a large margin in three out of four cases and had equal performance on the fourth, which had by far the least number of classes.

The \GraphMLP model without a classical message-passing pipeline is the best GNN model in three of the four cases.
The GNNs performed well for the \mclasscol{} graph summary model.
The ranking of the GNN models was not consistent across graph summary models, but the differences between them were rather small.
A classical $2$-layer MLP comes close to the results of the GNNs, being within $3\%$ of the best graph model.

In our experiments, we noticed that the accuracy of the neural network models decreases with increasing numbers of equivalence classes. 
The performance of our non-neural baseline, the Bloom filter, seemed to be independent of the number of classes.

\section*{Acknowledgements}
We thank the data science paper reading club of Ulm University and especially Lukas Galke for the input on implementation details on graph neural networks. Additional thanks go to Till Blume for the original proposal to investigate Bloom filters for graph summarization.

This research is the result of a Master module ``Project Data Science'' taught at Ulm University in 2020--2022.
The last two authors are the supervisors of the student group.

\bibliographystyle{IEEEtran}

\bibliography{Biblio.bib}

\newpage

\appendix

\subsection{Bloom Filter Hash Functions and Bits}
\label{sec:bloom-filter-details}
\autoref{tab:parameters_bloom} shows the number of hash functions and the bits of the Bloom filter array for the Bloom filter parameters that we used.
\begin{table}[htb]
    \centering
    \begin{tabular}{|l||r|r|}
        \hline
         \multicolumn{1}{|l||} {Bloom filter}  & \multicolumn{1}{l|} {$k$}  &\multicolumn{1}{l|} {$m$}\\
        \hline
        $n=4$, $p=10^{-1}$  & $3$   & $20$      \\
        $n=4$, $p=10^{-3}$  & $10$  & $58$      \\
        $n=4$, $p=10^{-7}$  & $23$  & $135$     \\
        $n=15$, $p=10^{-1}$ & $3$   & $72$      \\
        $n=15$, $p=10^{-3}$ & $10$  & $216$     \\
        $n=15$, $p=10^{-7}$ & $23$  & $504$     \\
        $n=60$, $p=10^{-1}$ & $3$   & $288$     \\
        $n=60$, $p=10^{-3}$ & $10$  & $863$     \\
        $n=60$, $p=10^{-7}$ & $23$  & $2{\small,}013$    \\
        \hline
    \end{tabular}
    \caption{The number of hash functions $k$ and the bits in the Bloom filter array $m$ according to our expected input items $n$ and false positive probability $p$.}
    \label{tab:parameters_bloom}
\end{table}
\label{sec:addStatsAppendixBloom}

\subsection{Optimal Hyperparameter Values}

A detailed summary of the hyperparameters that we used for the specific summary models and neural network models can be found in \autoref{tab:results_hyper}.

\begin{table}[!htb]
    \centering
    \begin{subtable}{0.45\textwidth}
        \centering
        \begin{tabularx}{1.\columnwidth}{|l||X|X|X|}
            \hline
                                & learning rate $\mu$   & hidden layer size & dropout   \\
            \hline
            MLP-2               & $0.1$                 & $64$              & $0.0$     \\
             GCN-2  & $0.1$                 & $64$              & $0.0$     \\
            GraphSAINT-2        & $0.1$                 & $64$              & $0.2$     \\
            GraphSAGE-2         & $0.1$                 & $64$              & $0.2$     \\
            \hline
        \end{tabularx}
        \caption{Best hyperparameter values for \mattrcol{}}
        \label{tab:results_hyper_attr}
    \end{subtable}
    \hfill
    \begin{subtable}{0.45\textwidth}
        \centering
        \begin{tabularx}{1.\columnwidth}{|l||X|X|X|}
            \hline
                                & learning rate $\mu$   & hidden layer size & dropout   \\
            \hline
            MLP-2               & $0.1$                 & $1024$            & $0.0$     \\
            GCN-2  & $0.1$                 & $64$              & $0.5$     \\
            GraphSAINT-2        & $0.1$                 & $64$              & $0.2$     \\
            GraphSAGE-2         & $0.1$                 & $64$              & $0.5$     \\
            \hline
        \end{tabularx}
        \caption{Best hyperparameter values for \mclasscol{}}
        \label{tab:results_hyper_cc}
    \end{subtable}
    \begin{subtable}{0.45\textwidth}
        \centering
        \begin{tabularx}{1.\columnwidth}{|l||X|X|X|}
            \hline
                                & learning rate $\mu$   & hidden layer size & dropout   \\
            \hline
            MLP-2               & $0.1$                 & $64$              & $0.2$     \\
            GCN-2  & $0.1$                 & $64$              & $0.5$     \\
            GraphSAINT-2        & $0.1$                 & $32$              & $0.0$     \\
            GraphSAGE-2         & $0.1$                 & $64$              & $0.2$     \\
            \hline
        \end{tabularx}
        \caption{Best hyperparameter values for \mpropertytypecol{}}
        \label{tab:results_hyper_ptc}
    \end{subtable}
    \hfill
    \begin{subtable}{0.45\textwidth}
        \centering
        \begin{tabularx}{1.\columnwidth}{|l||X|X|X|}
            \hline
                                & learning rate $\mu$   & hidden layer size & dropout   \\
            \hline
            MLP-2               & $0.1$                 & $1024$            & $0.5$     \\
            GCN-2  & $0.1$                 & $64$              & $0.2$     \\
            GraphSAINT-2        & $0.1$                 & $32$              & $0.0$     \\
            GraphSAGE-2         & $0.1$                 & $64$              & $0.2$     \\
            \hline
        \end{tabularx}
        \caption{Best hyperparameter values for \mschemexused{}}
        \label{tab:results_hyper_schemex}
    \end{subtable}
    \begin{subtable}{.45\textwidth}
        \centering
        \begin{tabularx}{1.\columnwidth}{|l||X|X|X|X|X|}    
            \hline
                                & learning rate $\mu$   & hidden layer size & temperature $\tau$    & weighting coefficient $\alpha$    \\
            \hline
            \mattrcol{}         & $0.01$                & $256$             & $1.0$                 & $100$                             \\
            \mclasscol{}        & $0.01$                & $256$             & $1.0$                 & $100$                             \\
            \mpropertytypecol{} & $0.01$                & $256$             & $2.0$                 & $10$                              \\
            \mschemexused{}     & $0.01$                & $256$             & $2.0$                 & $100$                             \\
            \hline
        \end{tabularx}  
        \caption{Best hyperparameter values for the different summary models for \GraphMLP.}
        \label{tab:results_hyper_GraphMLP}
    \end{subtable}
    \caption{Best hyperparameter values for the different graph summary models and for \GraphMLP.}
    \label{tab:results_hyper}
\end{table}
\label{sec:addStatsAppendixHyper}

\subsection{Parameter Count}
\label{sec:parameter-count-main-experiments}

The parameter counts for the competing models are given in \autoref{tab:parameter}.
The parameter counts depend on the neural network, given graph summary model, and  the chosen hidden layer size.

\begin{table}[htb]
    \centering
    \begin{tabular}{|l||r|r|r|r|}
        \hline
         \multicolumn{1}{|l||} {}  & \multicolumn{1}{l|} {\mattrcol{} }  &\multicolumn{1}{l|} { \mclasscol{} } & \multicolumn{1}{l|} {\mpropertytypecol{} }  & \multicolumn{1}{l|} {\mschemexused{}}   \\
        \hline
        MLP-2               & $11{\small,}525{\small,}017$  & $15{\small,}968{\small,}832$  & $12{\small,}561{\small,}832$          & $34{\small,}149{\small,}257$      \\
        GCN-2  & $11{\small,}525{\small,}017$  & $998{\small,}592$     & $12{\small,}561{\small,}832$          & $2{\small,}151{\small,}497$       \\
        GraphSAINT-2        & $22{\small,}895{\small,}705$  & $2{\small,}004{\small,}800$   & $12{\small,}563{\small,}880$          & $2{\small,}153{\small,}545$       \\
        GraphSAGE-2         & $22{\small,}887{\small,}449$  & $1{\small,}996{\small,}544$   & $24{\small,}945{\small,}128$          & $4{\small,}284{\small,}746$       \\
        \GraphMLP-1          & $45{\small,}678{\small,}809$  & $4{\small,}058{\small,}944$   & $49{\small,}778{\small,}216$          & $-$               \\
        \GraphMLP-2          & $-$           & $-$           & $-$                   & $8{\small,}617{\small,}353$       \\
        \hline
    \end{tabular}
    \caption{Parameter count of MLP-2 and the different graph neural network models in PyTorch Geometric.}
    \label{tab:parameter}
\end{table}

\subsection{Generating Training Data for the Summary Models}
\label{sec:GeneratingTrainingDataForTheSummaryModels}

We apply the \autoref{alg:DataConversion} to generate data for training, validation, and test based on subgraphs using \textit{SubjectInformation} for each subject in the dictionary. 
Here, we take into account the different hop characteristics of the graph summary models. 

\begin{algorithm}[htb]
\small
	\caption{Data conversion from \textit{SubjectInformation} dictionary to subgraph GraphData} 
	\label{alg:DataConversion}
	\begin{algorithmic}[1]
	\Procedure{DataConversion}{$k\text{-folds}$: integer, \Mi{}: graph summary model}
    	\For {every $\text{fold}\;f$ in $k\text{-folds}$}
    	    \For {every subject vertex $s$ in $f$}
    	        \State GraphData $data \leftarrow \emptyset$
    	        \State $SI_s \leftarrow$ GET SubjectInformation of $s$
    	        \State Transform $SI_s$ into GraphData and append to $data$
    	        \If{\Mi{} requires 2-hop}
    	            \For{Every $(p,o)$ in $SI_s$.Edges}
    	                \State $SI_o \leftarrow$ GET SubjectInformation of $o$
    	                \State 
    	                append $SI_o$ to $data$
    	            \EndFor
    	        \EndIf
    	        \State Store $data$ to disk
    	    \EndFor
        \EndFor
    \EndProcedure
	\end{algorithmic} 
\end{algorithm}

\subsection{Optimal Hyperparameter Values for No-Singleton}
\label{sec:addStatsAppendixNSHyper}

\label{sec:addStatsAppendixNSClassDistr}
A detailed overview of the hyperparameters used for the no-singletons ablation study is in \autoref{tab:results_hyper_NS}.
\begin{table}[htb]
    \centering
    \begin{subtable}{0.45\textwidth}
        \centering
        \begin{tabularx}{1.\columnwidth}{|l||X|X|X|}
            \hline
                                & learning rate $\mu$   & hidden layer size & dropout   \\
            \hline
            MLP-2               & $0.1$                 & $1024$            & $0.5$     \\
            GCN-2  & $0.1$                 & $32$              & $0.2$     \\
            GraphSAINT-2        & $0.1$                 & $64$              & $0.2$     \\
            GraphSAGE-2         & $0.1$                 & $64$              & $0.0$     \\
            \hline
        \end{tabularx}
        \caption{Best hyperparameter values for \mattrcolns{} without singleton classes.}
        \label{tab:results_hyper_attr_NS}
    \end{subtable}
    \hfill
    \begin{subtable}{0.45\textwidth}
        \centering
        \begin{tabularx}{1.\columnwidth}{|l||X|X|X|}
            \hline
                                & learning rate $\mu$   & hidden layer size & dropout   \\
            \hline
            MLP-2               & $0.1$                 & $64$              & $0.5$     \\
            GCN-2  & $0.1$                 & $64$              & $0.0$     \\
            GraphSAINT-2        & $0.1$                 & $64$              & $0.2$     \\
            GraphSAGE-2         & $0.1$                 & $64$              & $0.0$     \\
            \hline
        \end{tabularx}
        \caption{Best hyperparameter values for \mclasscolns{} without singleton classes.}
        \label{tab:results_hyper_cc_NS}
    \end{subtable}
    \begin{subtable}{0.45\textwidth}
        \centering
        \begin{tabularx}{1.\columnwidth}{|l||X|X|X|}
            \hline
                                & learning rate $\mu$   & hidden layer size & dropout   \\
            \hline
            MLP-2               & $0.1$                 & $1024$            & $0.0$     \\
            GCN-2  & $0.1$                 & $64$              & $0.2$     \\
            GraphSAINT-2        & $0.1$                 & $32$              & $0.0$     \\
            GraphSAGE-2         & $0.1$                 & $64$              & $0.0$     \\
            \hline
        \end{tabularx}
        \caption{Best hyperparameter values for \mpropertytypecolns{} without singleton classes.}
        \label{tab:results_hyper_ptc_NS}
    \end{subtable}
    \hfill
    \begin{subtable}{.45\textwidth}
        \centering
        \begin{tabularx}{.99\columnwidth}{|l||X|X|X|X|X|}    
            \hline
                                & learning rate $\mu$   & hidden layer size & temperature $\tau$    & weighting coefficient $\alpha$    \\
            \hline
            \mattrcol{}         & $0.01$                & $256$             & $2.0$                 & $1.0$                             \\
            \mclasscol{}        & $0.01$                & $64$              & $1.0$                 & $10.0$                            \\
            \mpropertytypecol{} & $0.01$                & $256$             & $1.0$                 & $1.0$                             \\
            \hline
        \end{tabularx}  
        \caption{Best hyperparameter values for the different summary models for \GraphMLP without singleton classes.}
        \label{tab:results_hyper_GraphMLP_NS}
    \end{subtable}
    \caption{Best hyperparameter values for the different graph summary models after the removal of singleton classes.}
    \label{tab:results_hyper_NS}
\end{table}

\subsection{Parameter Count for No-Singleton}
\label{sec:addStatsAppendixNSParamCount}

\autoref{tab:parameter_singleton} shows the parameter count for the no-singleton ablation study.
\begin{table}[htb]
    \centering\begin{tabular}{|l||r|r|r|}
        \hline
         \multicolumn{1}{|l||} {}  & \multicolumn{1}{l|} {\mattrcolns{} }  &\multicolumn{1}{l|} { \mclasscolns{} } & \multicolumn{1}{l|} {\mpropertytypecolns{} } \\
        \hline
        MLP-2               & $61{\small,}706{\small,}382$      & $991{\small,}962$         & $66{\small,}340{\small,}407$          \\
        GCN-2  & $1{\small,}972{\small,}110$       & $991{\small,}962$         & $4{\small,}192{\small,}887$           \\
        GraphSAINT-2        & $7{\small,}761{\small,}038$       & $1{\small,}991{\small,}642$       & $4{\small,}194{\small,}935$           \\
        GraphSAGE-2         & $7{\small,}752{\small,}782$       & $1{\small,}983{\small,}386$       & $8{\small,}335{\small,}991$           \\
        \GraphMLP-1          & $15{\small,}526{\small,}798$      & $996{\small,}250$         & $16{\small,}688{\small,}695$          \\ 
        \hline
    \end{tabular}
    \caption{Parameter count of the different graph summary models for the no singleton ablation study.}
    \label{tab:parameter_singleton}
\end{table}

\subsection{Implementation Details}
\label{subsec:implementation}
For the GNNs, we used the already implemented functionality provided by the PyTorch library~\cite{NEURIPS2019_9015} and the PyTorch Geometric extension~\cite{fey2019graph} for graphs. For \GraphMLP we adapted the original PyTorch code, distributed by its authors, to fit our data representation of PyTorch Geometric.

To run our experiments we used machine 1 (CPU: AMD EPYC 7F32 3.89 GHz; 1.96TB RAM) and machine 2 (CPU: AMD EPYC 7302 3.297 GHz; GPU: 4x NVidia A100-SXM4-40GB; 504GB RAM).
Both machines were used for different tasks to utilize the most out of every machine. On machine 1 all the preprocessing to generate the data, the Bloom filter evaluation, creation of plots and statistics were done using the faster CPU cores. Since the higher parallelization capability of GPUs allow faster training and inference times we used the GPUs of machine 2 for the hyperparameter search, GNN training, and the cross-validation.

\subsection{Further Ablation Studies}
\label{sec:AblationStudy}

We provide further insights by ablations and more detailed analyses.
Section~\ref{subsec:ASNoSingletons} reports the impact of removing the singleton classes from the graphs. 
Below, we additionally investigate the differences occurring when using $1$-hop GNNs instead of $2$-hop. 
Finally, we widen our MLP to test the impact of a bigger hidden layer size, as suggested by Galke et al.~\cite{galke2021forget}.

\subsubsection{1-hop GNNs for 1-hop Graph Summaries}
\label{subsec:AS1hop}
In our main experiments, we only apply $2$-hop graph neural network models. Because of the smaller parameter count than their $1$-hop counterpart, the computational cost to train them is also lower.
The increase in size of GNN-1 models is caused by the missing hidden layer. 
But most of our summary models consider only the information contained in their direct neighbours, in other words they consider $1$-hop of information. By applying $1$-hop neural models to those summaries, we investigate if neural models with a matching hop-number show a better performance or any other behavior.

We investigate this by choosing the two best performing GNN-2 models per graph summary from \autoref{tab:results} to run a $10$-fold cross-validation as an ablation study on their $1$-hop counterparts.
For comparison, we also run this study with \GraphMLP-1.
For a $1$-hop GCN, we reduce \autoref{eqn:gcn} to
\begin{equation}
    \label{eqn:gcn1}
    f_\text{GCN}(\mX,\mA) = \softmax \big( \hat{\mA} \cdot ( \mX \mW^{(0)} ) \big)\,.
\end{equation}
Due to the substantially higher computational costs of these GNN-1 models we could not apply all summary models as described in our main experiment. 
We use a smaller portion of the dataset to reduce the computational requirements. This smaller dataset was built by using $25\%$ of the root vertices of the whole dataset and therefore we call this dataset \dyldosmall{}.
We use the \dyldosmall{} dataset for the graph summary models \mattrcolsmall{}, \mclasscolsmall{} and \mpropertytypecolsmall{}. The number of equivalence classes of these models can be seen in \autoref{tab:numberEquiClasses1hop}\extended{~and their class distributions in Section~\ref{sec:addStatsAppendix1HClassDistr}}. 
Also, we excluded \mschemex{} since it is a $2$-hop graph summary.

\begin{table}[htb]
    \centering
    \begin{tabular}{|l||r|r|r|}
        \hline
         \multicolumn{1}{|l||} {}  & \multicolumn{1}{l|} {\mattrcolsmall{} }  &\multicolumn{1}{l|} { \mclasscolsmall{} } & \multicolumn{1}{l|} {\mpropertytypecolsmall{} } \\
        \hline
        Number of Classes   & $60{\small,}024$          & $411$             & $66{\small,}184$                  \\
        \hline
    \end{tabular}
    \caption{Number of equivalence classes for \mattrcolsmall{}, \mclasscolsmall{} and \mpropertytypecolsmall{} in the \dyldosmall{} dataset.}
    \label{tab:numberEquiClasses1hop}
\end{table}

We use the learning rates, we found during the hyperparameter optimization, \autoref{subsec:hyperparameter}, of the GNN-2 models. 
We can reuse the hyperparameter values since the structure of the datasets is the same. 
\label{sec:addStatsAppendix1HParamCount}
The parameter count for the $1$-hop ablation study can be found in \autoref{tab:parameter_1-hop}.
We run the experiments with the Adam optimizer for $75$ epochs, except for GraphSAGE-1 and \mpropertytypecolsmall{}. 
For the latter, we use SGD and $400$ epochs, which was necessary because of the higher memory usage of the Adam optimizer.

\begin{table}[htb]
    \centering\begin{tabular}{|l||r|r|r|}
        \hline
         \multicolumn{1}{|l||} {}  & \multicolumn{1}{l|} {\mattrcolsmall{} }  &\multicolumn{1}{l|} { \mclasscolsmall{} } & \multicolumn{1}{l|} {\mpropertytypecolsmall{} } \\
        \hline
        GCN-1  & $901{\small,}440{\small,}432$     & $6{\small,}172{\small,}398$   & $993{\small,}951{\small,}312$         \\
        GraphSAGE-1         & $1{\small,}802{\small,}820{\small,}840$   & $12{\small,}344{\small,}385$  & $1{\small,}987{\small,}836{\small,}440$       \\
        \hline
    \end{tabular}
    \caption{Parameter count of the different graph summary models for the $1$-hop ablation study.}
    \label{tab:parameter_1-hop}
\end{table}

In \autoref{tab:results1hop}, the results of the models per graph summary are listed.
All Bloom filter results are within a range of $0.0018$. \mattrcolsmall{} and \mpropertytypecolsmall{} show a larger result range than \mclasscolsmall{}. In contrast to our main experiment it could not be observed that a higher $n$ and lower $p$ always lead to a higher accuracy, but the results only differ at the fourth decimal digit.
Overall the standard errors are lower compared to their $2$-hop counterparts. The GCN-1 results for \mattrcolsmall{} and \mpropertytypecolsmall{} have increased accuracy measures. For the graph summary \mclasscolsmall{}, the accuracy measures are ${\sim}2\%$ lower. 
The biggest difference in accuracy are ${\sim}10\%$ for GraphSAGE-1 and \mpropertytypecolsmall{}. 
\GraphMLP-1 is the top performing GNN model for \mattrcolsmall{} and \mpropertytypecolsmall{} and the worst performer for \mclasscolsmall{}, as in our main experiment. Also, for \mclasscolsmall{} and \mpropertytypecolsmall{} the \GraphMLP-1 models have a higher standard error. 

\begin{table*}[htb]
    \centering
    \begin{tabular}{|l||r|r|r|}
        \hline
         \multicolumn{1}{|l||} {}  & \multicolumn{1}{l|} {\mattrcolsmall{} }  &\multicolumn{1}{l|} { \mclasscolsmall{} } & \multicolumn{1}{l|} {\mpropertytypecolsmall{} } \\
        \hline
        Bloom filter ($n=4$, $p=10^{-7}$)   & $0.8566$              & $0.8566$              & $0.8229$                  \\
        Bloom filter ($n=15$, $p=10^{-7}$)  & $0.8582$              & $0.8562$              & $0.8244$                  \\
        Bloom filter ($n=60$, $p=10^{-7}$)  & $0.8581$              & $0.8561$              & $0.8247$                  \\
        \hline
        GCN-1                  & $0.6066 \pm 0.0042$   & $0.8315 \pm 0.0024$   & $0.5709 \pm 0.0017$       \\
        GraphSAGE-1                         & $0.6064 \pm 0.0032$   & $0.8340 \pm 0.0025$   & $0.4647 \pm 0.0036$       \\
        \hline
        \GraphMLP-1                          & $0.6202 \pm 0.0044$   & $0.7945 \pm 0.0102$   & $0.5796 \pm 0.0101$       \\
        \hline
    \end{tabular}
    \caption{Results for $10$-fold cross-validation of the $1$-hop message-passing GNNs and Graph-MLP-1 reported by mean accuracy with standard error on the test fold (higher accuracy, lower standard error better).
    For direct comparison, the Bloom filter results are shown, too.
    The $n$-values for Bloom filter are chosen based on the $75$th, $95$th and $99$th percentile of the node degree distribution.
    There was no cross-validation applied to the Bloom filter results for the same reason as stated in \autoref{sec:results}. 
}
    \label{tab:results1hop}
\end{table*}

We found that the $1$-hop GNN models do not improve the overall results. 
This is likely due to the massively increased parameter count of the models.

\subsubsection{Widened MLP}
\label{subsec:ASMLP}
In our last ablation study, we explore the MLP-baseline model by widening it to check if the accuracy increases. This is motivated by Galke et al.~\cite{galke2021forget}. 
For this, we take the MLP-2 model from our experiments and increase the hidden layer size to $2{\small,}048$ and $4{\small,}096$, respectively. 
\label{sec:addStatsAppendixWMLPParamCount}
\autoref{tab:parameter_mlp} shows the parameter count for the widened MLP ablation study.
\begin{table}[htb]
    \centering
    \begin{tabular}{|l||r|r|r|}
        \hline
         \multicolumn{1}{|l||} {}  & \multicolumn{1}{l|} {\mattrcol{} }  &\multicolumn{1}{l|} { \mclasscol{} } & \multicolumn{1}{l|} {\mpropertytypecol{} } \\
        \hline
        MLP-2(original) & $11{\small,}525{\small,}017$  & $15{\small,}968{\small,}832$  & $12{\small,}561{\small,}832$          \\
        \hline
        MLP-2-2048        & $363{\small,}762{\small,}393$ & $31{\small,}937{\small,}088$  & $396{\small,}445{\small,}992$         \\
        MLP-2-4096        & $727{\small,}362{\small,}265$ & $63{\small,}873{\small,}600$  & $792{\small,}713{\small,}512$         \\
        \hline
    \end{tabular}
    \caption{Parameter count of the different graph summary models for the widened MLP ablation study.}
    \label{tab:parameter_mlp}
\end{table}

The results can be seen in \autoref{tab:resultsOtherMLPHiddenLayers}.
The accuracy does slightly improve for $2{\small,}048$, but does not outperform the other graph-based models. 
However, for $4{\small,}096$ hidden nodes, the scores actually decrease for \mattrcol{} and \mpropertytypecol{}. 
The standard error also increases considerably for those graph summaries.

\begin{table}[htb]
    \centering
    \begin{tabular}{|l||r|r|r|}
        \hline
         \multicolumn{1}{|l||} {}  & \multicolumn{1}{l|} {\mattrcol{} }  &\multicolumn{1}{l|} { \mclasscol{} } & \multicolumn{1}{l|} {\mpropertytypecol{} } \\
        \hline
        MLP-2(original) & $0.6033 \pm 0.0048$   & $0.8496 \pm 0.0067$   & $0.5706 \pm 0.0030$   \\
        \hline
        MLP-2-2048        & $0.6052 \pm 0.0079$   & $0.8510 \pm 0.0040$   & $0.5902 \pm 0.0061$   \\
        MLP-2-4096        & $0.6012 \pm 0.0067$   & $0.8307 \pm 0.0128$   & $0.5713 \pm 0.0039$   \\
        \hline
    \end{tabular}
    \caption{Results for $10$-fold cross-validation for wider hidden layer sizes MLP (higher accuracy, lower standard error better). The model names are composed by adding to MLP-2 the respective number of hidden vertices. The parameter count for these models can be found in Section~\ref{sec:addStatsAppendixWMLPParamCount}.}
    \label{tab:resultsOtherMLPHiddenLayers}
\end{table}

Since this ablation study shows that widening the MLP-2s did not improve the accuracy scores significantly, we conclude that it is not possible to improve the MLPs any further. 
Also since \GraphMLP-1 with $\alpha=0$ corresponds to a normal MLP-2, the comparison of MLP and \GraphMLP (see \autoref{tab:results}) shows the importance of the contrastive loss function.

\subsection{Further Class Distributions}

\extended{
In \autoref{fig:class_distr_ms_whole_dataset} a comparison of the class distribution of \mschemex{} and \mschemexused{} can be seen.
\autoref{fig:ClassDistributionsNoSingleton} shows the class distribution of the summary models on the DyLDO dataset after removal of the singleton classes.
The class distribution of the summary models on \dyldosmall{} can be seen in \autoref{fig:class_distr_1hop_datasets}.

\begin{figure*}[htb]
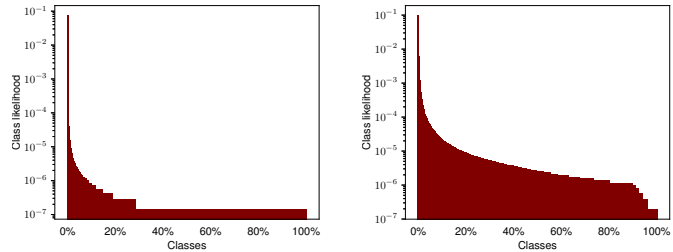

    \centering
    \begin{subfigure}{.25\textwidth}
        \begin{center}
            \resizebox{1.\columnwidth}{!}{\input{ressources/class_distribution_03.pgf}}
        \end{center}
        \caption{Class distribution \mschemex{} on the whole DyLDO dataset.}
    \end{subfigure} 
    \begin{subfigure}{.25\textwidth}
        \begin{center}
            \resizebox{1.\columnwidth}{!}{\input{ressources/class_distribution_03_min_support_5_1_500.pgf}}
        \end{center}
        \caption{Class distribution \mschemexused{} on the \dyldoschemex{} dataset.}
    \end{subfigure} 
    \caption{Class distribution of \mschemex{} (left) vs \mschemexused{} (right)}
    \label{fig:class_distr_ms_whole_dataset}
\end{figure*}
\label{sec:addStatsAppendixMsWhole}

\begin{figure*}[htb]
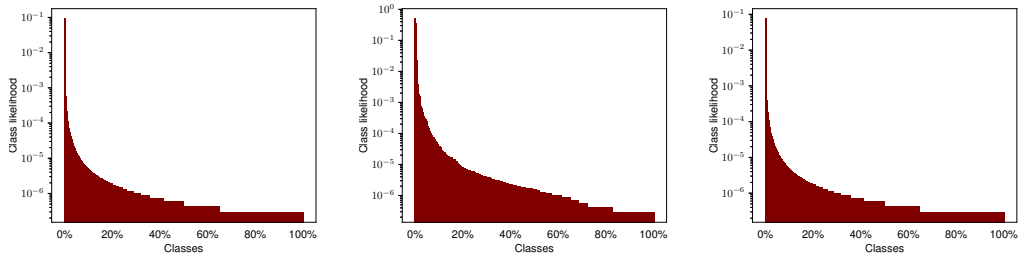

    \centering
    \begin{subfigure}{.25\textwidth}
        \begin{center}
            \resizebox{1.\columnwidth}{!}{\input{ressources/class_distribution_01_no_singleton.pgf}}
        \end{center}
        \caption{Class distribution \mattrcolns{}}
        \label{fig:class_distr_mac_NS}
    \end{subfigure} 
    \begin{subfigure}{.25\textwidth}
        \begin{center}
            \resizebox{1.\columnwidth}{!}{\input{ressources/class_distribution_02_no_singleton.pgf}}
        \end{center}
        \caption{Class distribution \mclasscolns{}}
        \label{fig:class_distr_mcc_NS}
    \end{subfigure}
    \begin{subfigure}{.25\textwidth}
        \begin{center}
            \resizebox{1.\columnwidth}{!}{\input{ressources/class_distribution_04_no_singleton.pgf}}
        \end{center}
        \caption{Class distribution \mpropertytypecolns{}}
        \label{fig:class_distr_mptc_NS}
    \end{subfigure} 
\caption{The class distributions of the different graph summary models on the DyLDO dataset after removal of the singleton classes. The x-axis shows the classes sorted in descending likelihood.}
\label{fig:ClassDistributionsNoSingleton}
\end{figure*}

\begin{figure*}[htb]
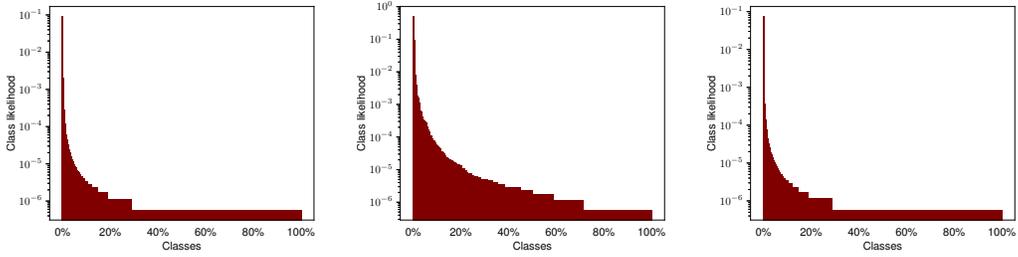

    \centering
    \begin{subfigure}{.25\textwidth}
        \begin{center}
            \resizebox{1.\columnwidth}{!}{\input{ressources/class_distribution_01_reduced_0_25.pgf}}
        \end{center}
        \caption{Class distribution \mattrcolsmall{}}
    \end{subfigure} 
    \begin{subfigure}{.25\textwidth}
        \begin{center}
            \resizebox{1.\columnwidth}{!}{\input{ressources/class_distribution_02_reduced_0_25.pgf}}
        \end{center}
        \caption{Class distribution \mclasscolsmall{}{}}
    \end{subfigure} 
    \begin{subfigure}{.25\textwidth}
        \begin{center}
            \resizebox{1.\columnwidth}{!}{\input{ressources/class_distribution_04_reduced_0_25.pgf}}
        \end{center}
        \caption{Class distribution \mpropertytypecolsmall{}}
    \end{subfigure} 
    \caption{Class distributions of \mattrcolsmall{} (top-left), \mclasscolsmall{} (top-right) and \mpropertytypecolsmall{} (bottom)}
    \label{fig:class_distr_1hop_datasets}
\end{figure*}
\label{sec:addStatsAppendix1HClassDistr}
}

\extended{
\subsection{Bloom Filter Impurity Results}

The Bloom filter impurity results for \dyldosmall{} can be seen in \autoref{tab:impurityBloomSmall}.
\begin{table*}[htb]
    \centering\begin{tabular}{|l||r|r|r|}
        \hline
         \multicolumn{1}{|l||} {}  & \multicolumn{1}{l|} {\mattrcolsmall{} }  &\multicolumn{1}{l|} { \mclasscolsmall{} } & \multicolumn{1}{l|} {\mpropertytypecolsmall{} } \\
        \hline
        $n=4$, $p=10^{-7}$      & $0.0502$          & $0.0604$          & $0.0626$                  \\
        $n=15$, $p=10^{-7}$     & $0.0498$          & $0.0606$          & $0.0622$                  \\
        $n=60$, $p=10^{-7}$     & $0.0498$          & $0.0606$          & $0.0621$                  \\
        \hline
    \end{tabular}
    \caption{Impurity measure for Bloom filter using the Gini-index on \dyldosmall{} with expected input items $n$ and false positive probability $p$ (lower impurity better).}
    \label{tab:impurityBloomSmall}
\end{table*}
\label{sec:addStatsAppendix1HBloom}
}

\extended{
\subsection{Detailed Discussion of Simple-GCN}
\label{sec:appendixSGCN}

We also considered using the SGC model by Wu et al.~\cite{DBLP:journals/corr/abs-1902-07153}.  
Their goal is to reduce the complexity of GCN by removing non-linearities and by condensing the weight matrices of single layers into a combined one. 
This results in the following simplification of \autoref{eqn:gcn}
\begin{equation}
\label{eqn:sgcn}
    f_\text{SGC}(\mX,\mA) = \softmax\Big( \hat{\mA}^{k} \cdot ( \mX \mTheta ) \Big)
\end{equation} 
with $\mTheta = \mW^{(0)} \mW^{(1)} \ldots \mW^{(k)} $ for $k$-hops.
The $1$-hop SGC is equivalent to the $1$-hop GCN (see \autoref{eqn:gcn1} and \autoref{eqn:sgcn} for $k=1$ and 1-layer GCN).

In a few pre-experiments on the DyLDO dataset with only $30M$ edges, we call this dataset \dyldothirty{}, SGC scored considerably lower than the other GNNs (see \autoref{tab:resultsSGCN}), and was also the biggest model, which pushed us to the limits of our GPU memory. In our main experiments on the DyLDO dataset, the number of classes increased for \mattrcol{}, \mpropertytypecol{} and \mschemex{} which in turn increased the number of parameters for our models. Now the SGC model exceeded the GPU memory. Because it also had a low performance in our pre-experiments, we decided to exclude this from our main paper.

For our experiments we report the results for the summary models on \dyldothirty{} as \mattrcolthirty{}, \mclasscolthirty{} and \mpropertytypecolthirty{}.
\begin{table*}[htb]
    \centering
    \begin{tabular}{|l||r|r|r|}
        \hline
         \multicolumn{1}{|l||} {}  & \multicolumn{1}{l|} {\mattrcolthirty{} }  &\multicolumn{1}{l|} { \mclasscolthirty{} } & \multicolumn{1}{l|} {\mpropertytypecolthirty{} } \\
        \hline
        SGC-2 & $0.2079 \pm 0.0027$   & $0.2542 \pm 0.0036$   & $0.0818 \pm 0.0012$       \\
        GraphSAGE-2 & $0.6929 \pm 0.0044$   & $0.8771 \pm 0.0023$   & $0.6389 \pm 0.0041$       \\
        \hline
    \end{tabular}
    \caption{$10$-fold cross-validation results for SGC (higher accuracy, lower standard error better).}
    \label{tab:resultsSGCN}
\end{table*}
}

\end{document}